\documentclass{article}
\usepackage{iclr2020_conference,times}
\pdfoutput=1

\usepackage{amsmath,amsfonts,bm}

\def\eqref#1{equation~\ref{#1}}

\def\1{\bm{1}}

\DeclareMathAlphabet{\mathsfit}{\encodingdefault}{\sfdefault}{m}{sl}
\SetMathAlphabet{\mathsfit}{bold}{\encodingdefault}{\sfdefault}{bx}{n}

\usepackage{hyperref}
\usepackage{url}

\usepackage{algorithm}
\usepackage{algorithmic}
\usepackage{graphicx}
\usepackage{amsmath}
\usepackage{amsfonts}
\usepackage{mathtools}
\usepackage{xspace}
\usepackage{multirow}
\usepackage{subcaption}
\usepackage[export]{adjustbox}
\usepackage{verbatim}
\newcommand{\exrec}{ExRec\xspace}
\newcommand{\person}{\mathcal{P}}
\newcommand{\task}{\mathcal{T}}
\newcommand{\data}{\mathcal{D}}
\newcommand{\loss}{\mathcal{L}}

\newcommand{\learner}{f}
\newcommand{\classes}{\mathcal{C}}
\newcommand{\lossi}{\loss_{\person_i}}
\newcommand{\learnerfr}{\learner_{\theta_f,\theta_r}}
\newcommand{\mex}{MEx\xspace}
\def\sfb{\textsc{selfBACK}\xspace}
\def\pamap{\textsc{PAMAP2}\xspace}
\newcommand{\mexacw}{MEx$_{ACW}$\xspace}
\newcommand{\mexact}{MEx$_{ACT}$\xspace}
\newcommand{\mexpm}{MEx$_{PM}$\xspace}
\newcommand{\mexdc}{MEx$_{DC}$\xspace}
\newcommand{\sbw}{SB$_{W}$\xspace}
\newcommand{\sbt}{SB$_{T}$\xspace}

\newcommand{\pamapa}{PMP$_{A}$\xspace}
\newcommand{\pamaph}{PMP$_{H}$\xspace}
\newcommand{\pamapc}{PMP$_{C}$\xspace}

\newcommand{\row}[1]{\renewcommand{\arraystretch}{#1}}

\title{Learning-to-Learn Personalised Human Activity Recognition Models}

\author{Anjana Wijekoon, Nirmalie Wiratunga \thanks{This work is part funded by \sfb{} which has received funding from the European Union’s Horizon 2020 research and innovation programme under grant agreement No 689043.} \\
School of Computing and Digital Media\\
Robert Gordon University\\
Aberdeen, UK \\
\texttt{\{a.wijekoon,n.wiratunga\}@rgu.ac.uk} \\
}

\iclrfinalcopy
\begin{document}

\maketitle

\begin{abstract}
Human Activity Recognition~(HAR) is the classification of human movement, captured using one or more sensors either as wearables or embedded in the environment~(e.g. depth cameras, pressure mats).
State-of-the-art methods of HAR rely on having access to a considerable amount of labelled data to train deep architectures with many train-able parameters.
This becomes prohibitive when tasked with creating models that are sensitive to personal nuances in human movement, explicitly present when performing exercises.
In addition, it is not possible to collect training data to cover all possible subjects in the target population.
Accordingly, learning personalised models with few data remains an interesting challenge for HAR research.
We present a meta-learning methodology for learning to learn personalised HAR models for HAR; with the expectation that the end-user need only provides a few labelled data but can benefit from the rapid adaptation of a generic meta-model.
We introduce two algorithms, Personalised MAML and Personalised Relation Networks inspired by existing Meta-Learning algorithms but optimised for learning HAR models that are adaptable to any person in health and well-being applications.
A comparative study shows significant performance improvements against the state-of-the-art Deep Learning algorithms and the Few-shot Meta-Learning algorithms in multiple HAR domains.
\end{abstract}

\section{Introduction}

Machine Learning research in Human Activity Recognition~(HAR) has a wide range of high impact applications in gait recognition, fall detection, orthopaedic rehabilitation and general fitness monitoring.
HAR involves the task of learning reasoning models to recognise activities from human movements inferred from streams of sensor data.
A data instance, in a HAR dataset, contains sensor data streams collected from individuals (i.e. person data). 
Unavoidably, sensor streams capture personal traits and nuances in some activity domains more than others. Typically with activities that involve greater degrees of freedom. 
Learning a single reasoning model to recognise the set of activity classes in a HAR task can be challenging because of the need for personalisation. 

We propose it is more intuitive to treat a ``person-activity'' pair as the class label.
This means for example if two people were to perform the same movement activity, they should be treated as separate classes because they are executed by 2 different people. 
Accordingly, each person's data can be viewed as a dataset in its own right, and the HAR task involves learning a reasoning model for the person.
Learning from only specific persons' data has shown significant performance improvements in early research with both supervised learning and active learning methods~\cite{tapia2007real,longstaff2010improving}. 
But these methods require considerable amounts of data obtained from the end-user, periodical end-user involvement and model re-training. 
In addition, current state-of-the-art Deep Learning algorithms require a large number of labelled data instances to avoid under-fitting.

Here we adopt the ``person-activity'' classes idea but attempt to learn with a limited number of data instances per class.
This can be viewed as a Few-shot classification scenario~\cite{vinyals2016matching,snell2017prototypical} where the aim is to learn with one or few data instances and is commonly evaluated with datasets such as Omniglot (1623 classes) and MiniImagenet (100 classes), but only limited by the number of data instances available for each class. 
Meta-Learning is arguably the state-of-the-art in Few-shot classification for image recognition~\cite{finn2017model,nichol2018first}. 
In a nutshell, Meta-Learning is described as learning-to-learn, where a wide range of tasks abstract their learning to a meta-model, such that, it is transferable to any unseen task. 
Meta-Learning algorithms such as MAML~\cite{finn2017model}, Relation Networks~(RN)~\cite{sung2018learning} are grounded in theories of Metric Learning and parametric optimisation, and capable of learning generalised models, rapidly adaptable to new tasks with only a few instances of data. 

The concept of learning-to-learn aligns well with personalisation where modelling a person can be viewed as a single task; whereby the meta-model must help learn a model that is rapidly adaptable to a new person. 
We propose \textit{Personalised Meta-Learning} to create personalised models, by leveraging a small amount of sensing data~(i.e. calibration data) extracted from a person. 
Accordingly, in this paper we make the following contributions, 
\begin{enumerate}
\item formalise Personalised Meta-Learning and propose two Personalised Meta-Learning Algorithms, Personalised MAML and Personalised RN;
\item perform a comparative evaluation with 9 HAR datasets representing a wide range of activity domains to evidence the utility of Personalised Meta-Learning algorithms over conventional Deep Learning and Few-shot Meta-Learning algorithms;
\item analyse train and test results of personalised vs. conventional meta-learners to understand how personalisation enhanced meta-learners that are able to adapt a generalised model at deployment; and
\item present an exploratory study to explore hyper-parameter selection of personalised meta-learners.
\end{enumerate}

Overall, we observe that the performance improvements achieved by Personalised Meta-Learning is possible using simple parametric models with limited number of trainable parameters that only require a limited amount of labelled data compared to conventional DL models. 
The rest of the paper is organised as follows: Section~\ref{sec:related} explore past research, challenges in the areas of Personalised HAR. 
Section~\ref{sec:meta-learning} introduces the state-of-the art in Meta-Learning and our proposed approach and algorithms for Personalised Meta-Learning is presented in Section~\ref{sec:method}.
Next we present our comparative study, including datasets and evaluation methodology in Section~\ref{sec:compare}. 
In Section~\ref{sec:pvsnp} we compare how personalisation improve the performance of Meta-Learners and in Section~\ref{sec:hyperpara} we explore a number of hyper-parameters for optimal performance of Personalised Meta-Learners. 
Finally a discussion on practical implication, limitations and planned future work are presented in Section~\ref{sec:discuss} and we present our conclusions in Section~\ref{sec:conc}.
\section{Related Work}
\label{sec:related}
Human Activity Recognition~(HAR) is an active research challenge, where Deep Learning~(DL) methods claim the state-of-the-art in many application domains~\cite{wijekoon2019mex,wang2019deep,ordonez2016deep,yao2017deepsense}. 
Learning a generalised reasoning model adaptable to many user groups is a unique transfer learning challenge in the HAR domain. 
Sensors capture many personal nuances, that are most prominent in application domains such as Exercises or Activities of Daily Living~(ADL), leading to poor performance. Given access to large quantities of end-user data, early research has achieved improved performance by learning personal models~\cite{tapia2007real,berchtold2010actiserv}. Follow on work attempts to reduce the burden on end-user, by adopting semi-supervised~\cite{longstaff2010improving,miu2015bootstrapping}, active learning~\cite{longstaff2010improving} and multi-task~\cite{sun2012large} methods that rely on periodical model re-training and continuous user involvement post-deployment.

Recent advancements in Few-shot Learning are adopted as an approach to personalisation in Personalised Matching Networks ($MN^p$)~\cite{wijekoon2020knowledge,vinyals2016matching}. $MN^p$ learns a parametric model, that is learning to match, leveraging a few data instances from the same user. At deployment, the network successfully transfers the learning to new users given only a few labelled data instances for matching, obtained through one-time micro-interactions. 
This approach avoids post-deployment re-training and only require a few data instances from the end-user. 
While this method achieves significant performance improvement over conventional methods in HAR, we believe moderated post-deployment re-training can be beneficial for improving personalisation.

Meta-Learning is an interesting approach that will facilitate Few-shot Learning and post-deployment re-training for adaptation.
Meta-Learning or ``Learning-to-learn'' is the learning of a generalised classification model that is transferable to new learning tasks with only a few instances of labelled data. In recent research it is interpreted and implemented mainly in three approaches; firstly, ``learning to match'' approach implemented by Relation Networks~(RN)~\cite{sung2018learning}; secondly, model-specific approach like SNAIL~\cite{mishra2017simple}; and finally, optimisation based algorithms such as MAML~\cite{finn2017model} and Reptile~\cite{nichol2018first}.

MAML, including its variants such as First-Order MAML~\cite{finn2017model} and Reptile~\cite{nichol2018first}, is an optimisation based Meta-Learning algorithm, learns a generalised model rapidly adaptable to any new task. Notably, these models are model-agnostic, which increases the adaptability in new domains where different feature-representation are preferred. In contrast, there exist, model-specific Meta-Learning algorithms, such as SNAIL~\cite{mishra2017simple} and MANN~\cite{santoro2016meta}, where Meta-Learning is achieved using specific neural network constructs such as LSTM and Neural Turing Machine~\cite{graves2014neural}.  Model-agnostic methods are preferred in a HAR setting, where heterogeneous sensor modalities or modality combinations may require specific feature representation learning methods. 
Relation Networks~(RN)~\cite{sung2018learning} are ``learning to match'' by learning similarity relationships. 
While there are significant commonalities between the Few-shot Learning algorithm MN~\cite{vinyals2016matching} and RN, it is not limited by a distance metric for similarity calculation.
The parametric model for similarity learning in RN, enhances the learning from a Few-shot classifier to a generalised meta-model. 

In comparison to model-agnostic and model specific methods, RN is also model-agnostic where the architecture is modularised~\cite{sung2018learning} such that the feature representation learning can be adapted to suit the HAR task. In contrast to model-agnostic and model-specific methods, RN has the potential to perform Open-ended HAR, by modelling the classification task as a matching task~(i.e. no softmax layer with fixed class length) similar to Open-ended Matching Networks~\cite{wijekoon2020knowledge}.
In this paper we explore personalisation of Meta-Learning for HAR with two algorithms, MAML and RN. Personalised Meta-Learners will only require limited interaction with the end-user to obtain few data instances and facilitate optional post-deployment re-training, which are both essential features for personalised HAR.

\section{Meta-Learning}
\label{sec:meta-learning}

\begin{figure}[ht]
\centering
\includegraphics[width=0.65\textwidth]{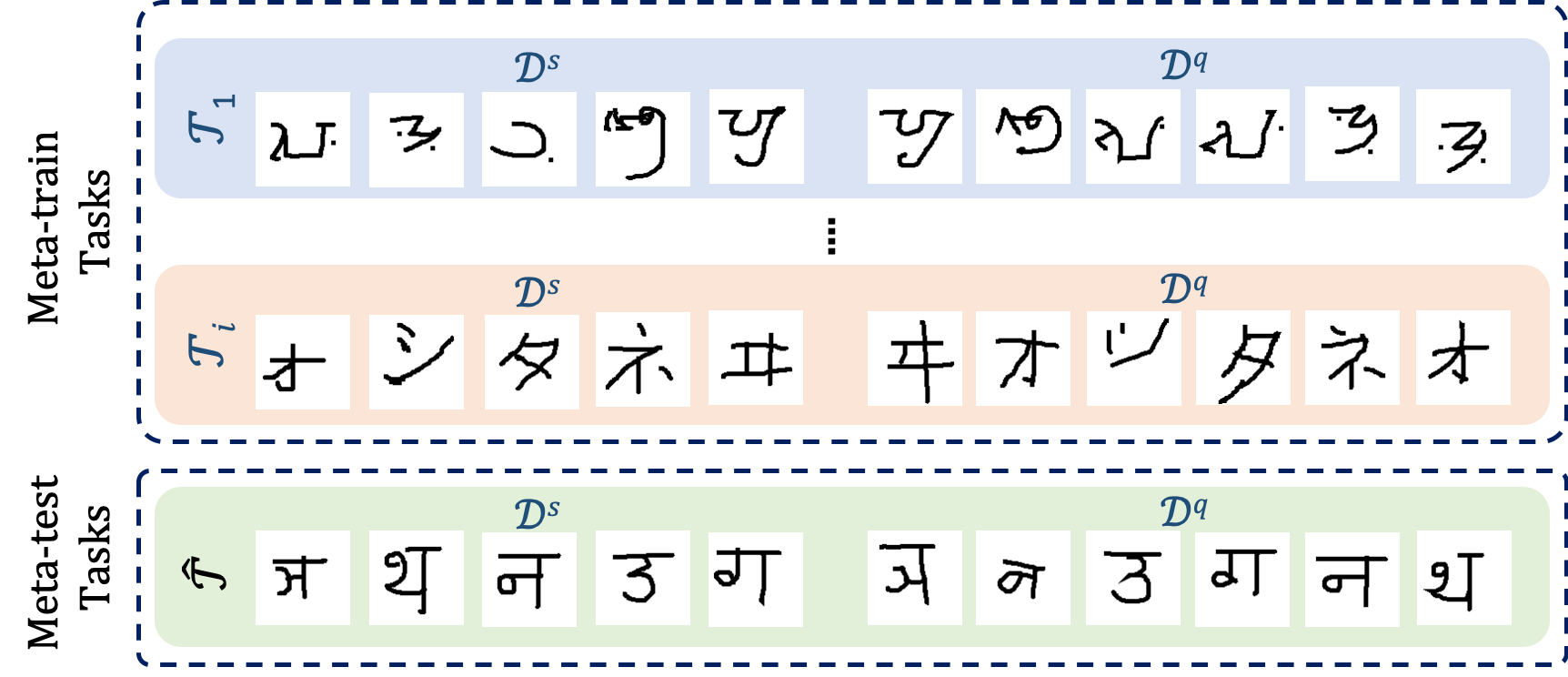}
\caption{Meta-Learning Tasks with Omniglot Dataset}
\label{fig:omniglot-meta}
\end{figure}
Meta-Learning learns a meta-model, $\theta$, trained over many tasks, where a task is equivalent to a ``data instance'' or "labelled example" in conventional Machine Learning~(ML). 
In practice, meta-learning is implemented as an optimisation over the set of tasks to learn a generalised model (i.e. meta-model $\theta$) that can be rapidly transferred to new, seen or unseen tasks.
The concept of Meta-Learning is implemented in many branches of Machine Learning~(ML), such as Few-shot learning, Reinforcement Learning and Regression, and here our focus is Few-shot Learning. 

In Few-shot classification, Meta-Learning can be seen as optimisation of a parametric model over many few-shot tasks~(i.e. meta-train).
More formally, a task, $\task$ is a few-shot learning problem with a set of train and test data instances, referred to as the ``support set'', $\data^s$, and the ``query set'', $\data^q$. 
The number of instances in a support set is ,$K^s \times |\classes|$, where $\classes$ is the set of classes and $K^s$ is the number of representatives from each class in the support set(often also referred to as a the shots in k-shot learning). 
For example, in Figure~\ref{fig:omniglot-meta}, a task support set contains distinct digits each of which forms a class (here $\classes$ = 5 with $K^s$ = 1). Each task contains an equal number of classes but not necessarily the same sub set of classes. 
Typically the query set, $\data^q$, has no overlap with the support set, $\data^s$ similar to a train/test split in supervised learning; and unlike the support set, composition of the query set need not be constrained to represent all $\classes$.

Once the meta-model is trained using the meta-train tasks, it is tested using the meta-test tasks. 
A meta-test task, $\hat{\task}$, has a similar composition to a meta-train task, in that it also has a support set and a query set. 
Unlike traditional classifier testing; with meta-testing, we use the support set in conjunction with the trained meta-model to classify instances in the query sets. 
For few shot learning there are two common meta-learning approaches: the adaptation optimised algorithms such as MAML~\cite{finn2017model} and Reptile~\cite{nichol2018first}; and the similarity optimised learning of Relation Networks~\cite{sung2018learning}. 

\begin{figure}[ht]
\centering
\includegraphics[width=0.7\textwidth]{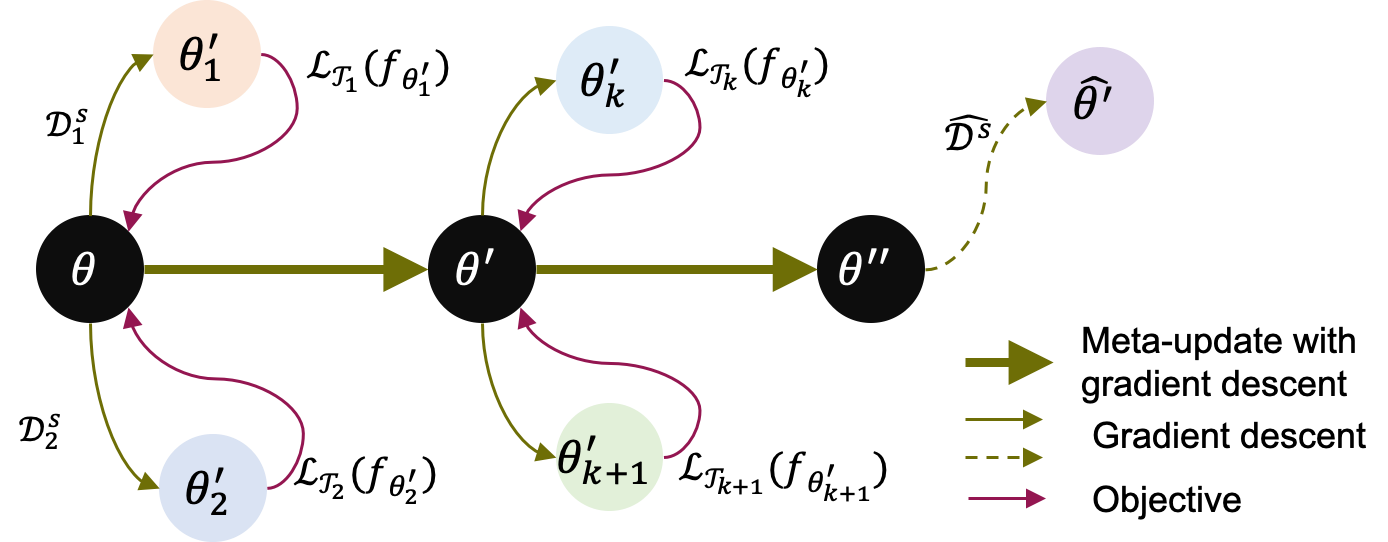}
\caption{Model-Agnostic Meta-Learning}
\label{fig:maml}
\end{figure}

MAML~\cite{finn2017model} is a versatile Meta-Learning algorithm applicable to any neural network model optimised with Gradient Descent~(GD). 
Adaptation optimised Meta-Learner MAML is illustrated in Figure~\ref{fig:maml}. 
In each training iteration, a set of tasks are sampled and each task, $\task_i$ optimises its task-model, $\theta_i$ with the $\data^s$ using one or few steps of GD referred to as gradient steps~($gs$). 
The meta-modal $\theta$ is then optimised using GD with the losses computed by the optimised task-models $\theta_i'$s against their respective $\data^q$s, referred to as the Meta-update.
This process is repeated with many task samples, to learn a generic model prototype $\theta$ that can be rapidly adapted to a new task. 
A task, $\hat{\task}$, not seen during training, uses its support set, $\data^s$ to train a parametric model $\hat{\theta}$, initialised by the meta-model $\theta$, for few gradient steps referred to as, meta-gradient-steps, $meta\_gs$. 
Thereafter, the adapted, $\hat{\theta}$ is used to classify instances in its query set, $\data^q$.

\begin{figure}[ht]
\centering
\includegraphics[width=0.75\textwidth]{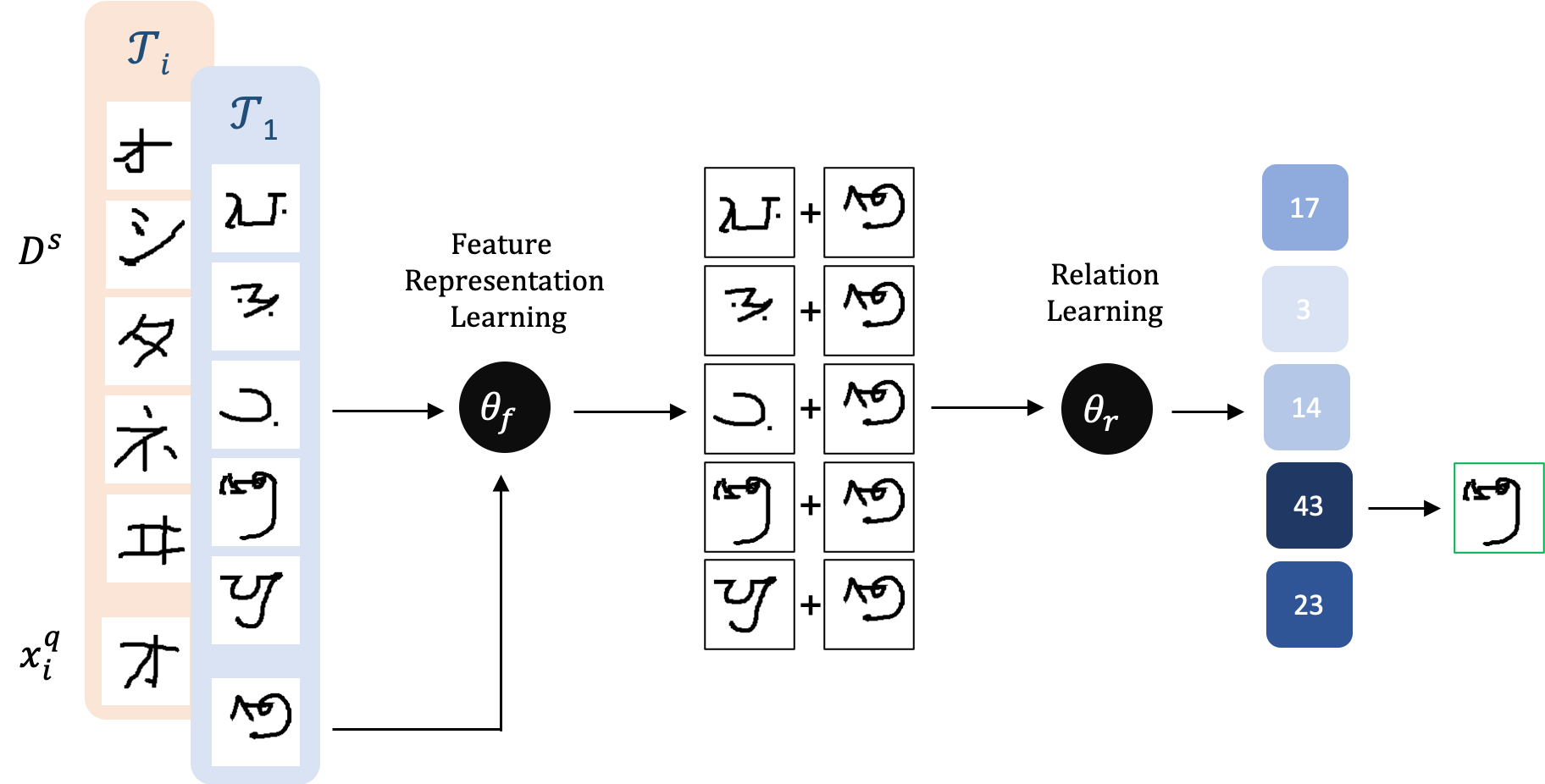}
\caption{Relation Network}
\label{fig:rn}
\end{figure}
Relation Network~(RN)~\cite{sung2018learning} is a Few-shot Meta-Learning algorithm that ``learns-to-match'' or learns a non-linear similarity function for matching.
RN has a similar goal to other Meta-Learners, of generalising over many tasks. The meta-task design for RN is as described in Figure~\ref{fig:omniglot-meta}. However with RN, each train-task consists of a number of meta-training instances, created by combining each $x_i^q$ from, $\data^q$ with the support set $\data^s$. 
During training, RN learns to match each instance, $x_i^q$ to a matching instance in $\data^s$. As illustrated in Figure~\ref{fig:rn}, the $\theta_f$ learns a feature representation for each instance; next each $x_i^s$, in $\data^s$ is paired with the $x_i^q$. The relation network then learns to estimate the similarity of the paired instances. 
The network is optimised end to end such that we learn a feature representation model, $\theta_f$, and a relation model, $\theta_r$, that collectively maximises the similarity between pairs that belong to the same class.
A meta-test task, $\hat{\task}$, not seen during training, can use the RN to match a query instance in $\data^q$ to an instance in it's support set $\data^s$, and therein use the class of the matched support instance as the predicted class label.

\section{Methods}
\label{sec:method}

Given a dataset, $\mathcal{D}$, Human Activity Recognition~(HAR), like any other supervised learning problem, is the learning of the feature mapping $\theta$ between data instances, $x$, and activity classes, $y$, where $y$ is from the set of activity classes, $\classes$. 
\begin{equation}
y = \theta(x) \text{ where } y \in \mathcal{C}
\end{equation}
In comparison to image or text classification, with HAR, each data instance in $\mathcal{D}$ belongs to a person, $p$. Given the set of data instances obtained from person $p$ is $\mathcal{D}^{p}$, $\mathcal{D}$ is the collection of data instances from the population $\mathcal{P}$~(Equation~\ref{eq:pd}). As before, all data instances in $\mathcal{D}^{p}$ will belong to a class in $\classes$, except for special tasks such as open-ended HAR where $\classes$ is not fully specified at training time. 
\begin{equation}
\mathcal{D} = \{\mathcal{D}^{p} \mid p\in\mathcal{P}\} \text{ where } \mathcal{D}^p = \{(x,y) \mid y\in\classes\}\\
\label{eq:pd}
\end{equation}

\subsection{Personalised Meta-Learning for HAR}
\label{sec:pml}
\begin{figure}[ht]
\centering
\includegraphics[width=0.65\textwidth]{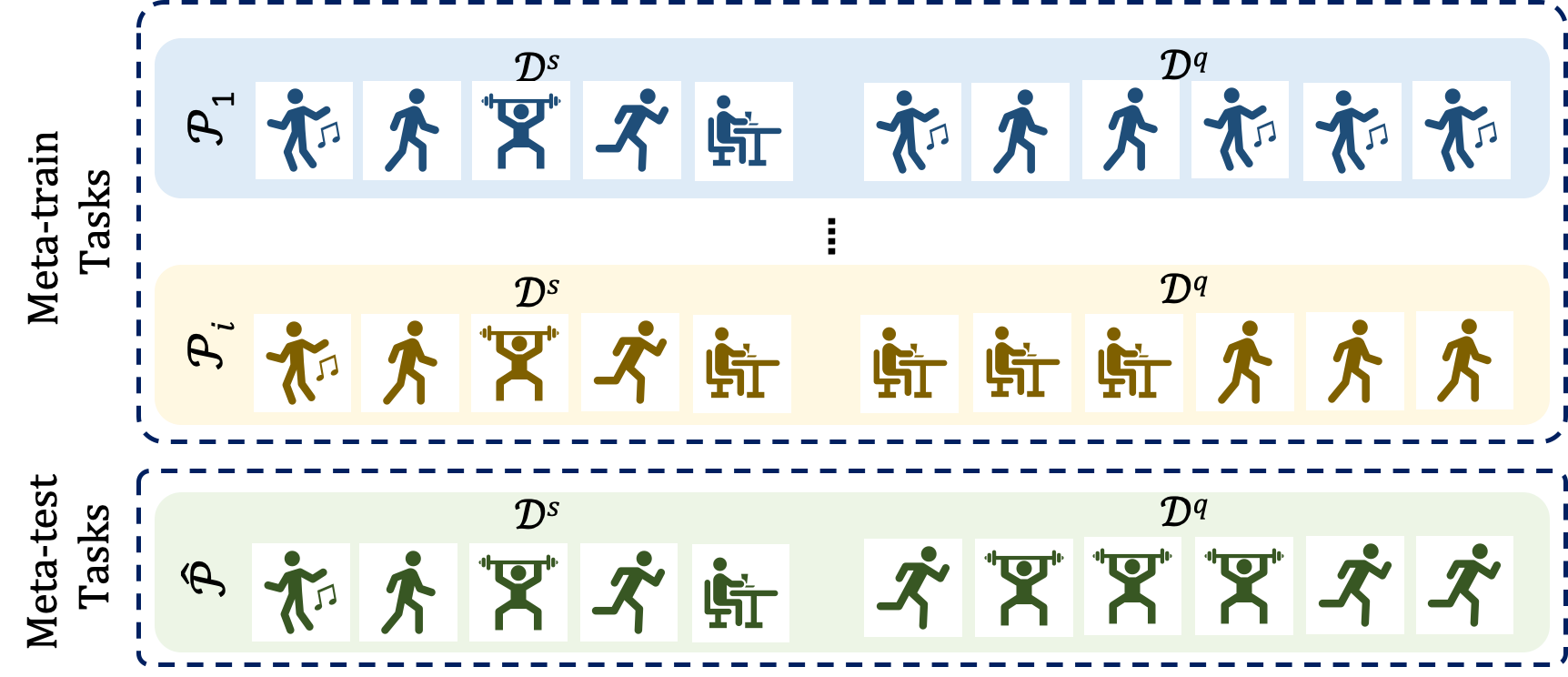}
\caption{Personalised Meta-Learning Tasks design for HAR}
\label{fig:har-meta}
\end{figure}
Personalised Meta-Learning for Human Activity Recognition~(HAR) can be seen as the learning of a meta-model $\theta$ from a population $\mathcal{P}$ while treating activity recognition for a person as an independent task. 
We propose the task design in Figure~\ref{fig:har-meta} for Personalised Meta-Learning. 
Given a dataset $\data$, of population $\person$, we create tasks such that, each ``person-task'', $\person_i$, only contain data from a specific person, $p$. 
We randomly select a $K^s \times |\classes|$ number of labelled data instances from person $p$ stratified across activity classes, $\classes$, such that there are $K^s$ amount of representatives for each class. We follow a similar approach when selecting a query set, $\data^q$, for $\person_i$. 
Given that existing HAR dataset are not strictly Few-shot Learning datasets, there can be a few to many data instances available to be sampled for the query set, $\data^q$. 
In comparison to Meta-Learning task design~(Section~\ref{sec:meta-learning}), each ``person-task'' is learning to classify the set of ''person-activity'' class labels. 

A dataset is split for training and testing using a person-aware evaluation methodology, such as Leave-One-Person-Out or Persons Hold-out where 1 or few persons form the meta-test person-tasks and the rest form the meta-train person-tasks. 
At test time, the test person, $\hat{p}$, provides a few seconds of data for each activity class while being recorded by recommended sensor modalities, which forms the support set, $\data^s$, of the person-task, $\hat{\person}$. Thereafter, the meta-model, in conjunction with the support set, predicts the class label for each query data instance, $x^q_i$, in $\data^q$.

It is noteworthy that, contrary to conventional Meta-Learning, all personal models and the meta-model are learning to classify the same set of activity classes $\classes$, but of different persons~(i.e. ``person-activity''). Therefore, it is seen as a Few-shot Meta-Learning classification problem with a $|\classes| \times |\person|$ number of classes. 
Personalised Meta-Learning is a methodology adaptable with any Meta-Learning algorithm to perform personalised HAR, and next we show how with two Meta-Learning algorithms, MAML and RN. 

\subsection{Personalised MAML}
\label{sec:pmaml}

Personalised MAML~($MAML^p$) for HAR is adaptation optimised to learn the generic model prototype~(i.e. meta-model), $\theta$, such that it is adaptable to any new person encountered at test time. 
Task design for $MAML^p$ follow the similar process described in Section~\ref{sec:pml} where we select a support set, $\data^s$, and a query set $\data^q$ for the person-task $\person_i$. 
The number of instance in the support set  $|\data^s|$, determines the number of instances that need to be requested from a new person, $\hat{p}$, during testing. Thus, we keep $K^s$ small, similar to a few-shot learning scenario. 
We use all remaining data instances from each class, in $\data^q$. More formally, given there are $K$ instances per ``person-activity'', $K^q = (K - K^s)$ and $\mid\data^q\mid = K^q \times \mid\classes\mid$.  
The meta-model learning, is influenced by the loss generated by the $\data^q$ of each person-task. Therefore we evaluate $MAML^p$ for a range of $K^q$ values in an exploratory study. We present the training of Personalised MAML in Algorithm~\ref{algo:pmaml_train}.

A meta-test person $\hat{p}$ provide $K^s$ instances per class forming the support set $\data^s$, that is used to train the personalised classification model, $\hat{\theta}$, initialised by the meta-model $\theta$, for $\hat{p}$. This process can be seen as the personalised model adaptation from the meta-model $\theta$. Thereafter, the class label is predicted for each incoming test data instances using $\hat{\theta}$ as in Algorithm~\ref{algo:pmaml_test}.

Personalised MAML is model-agnostic, with the opportunity to use a diverse range neural networks for feature representation learning. This is advantages for HAR applications where heterogeneous sensor modalities or modality combinations are used. 
We note that we refer to First-Order MAML~\cite{finn2017model} when implementing Personalised MAML, which is computationally less intensive, yet achieves comparable performances in comparison to MAML~\cite{finn2017model}.

\begin{minipage}{0.55\textwidth}
\begin{algorithm}[H]
\caption{Personalised MAML Training}
\label{algo:pmaml_train}
\begin{algorithmic}[1]
\REQUIRE $p(\person)$: HAR dataset; distribution over persons
\REQUIRE $\alpha$, $\beta$, $n$, $gs$, $meta\_gs$: step sizes, batch size and gradient-steps hyper-parameters
\STATE randomly initialise $\theta$
\WHILE{not done}
\STATE Sample $n$ persons $\person_i \sim p(\person)$
  \FORALL{$\person_i$}
    \STATE $\data^s = \{(x,y) \in \person_i : |\data^s| = K^s \times |\classes|$\}
    \FOR{$i=0$ \TO $gs$}
        \STATE Evaluate $\nabla_\theta \lossi(\learner_\theta)$ w.r.t. $\data^s$
        \STATE Compute adapted parameters with gradient descent: $\theta_i'=\theta-\alpha \nabla_\theta  \lossi(  \learner_\theta )$
    \ENDFOR
    \STATE $\data^q = \{(x,y) \in \person_i : \data^s \cap \data^q = \emptyset, |\data^q| = K^q \times |\classes|$\}
    \STATE Evaluate $\lossi(\learner_{\theta_i'})$ w.r.t $\data^q$
  \ENDFOR
  \STATE Meta-update: $\theta \leftarrow \theta - \beta \nabla_\theta \sum_{\person_i \sim p(\person)}   \lossi ( \learner_{\theta_i'})$ 
\ENDWHILE
\end{algorithmic}
\end{algorithm}
\end{minipage}
\hfill
\begin{minipage}{0.40\textwidth}
\begin{algorithm}[H]
\caption{Personalised MAML Testing}
\label{algo:pmaml_test}
\begin{algorithmic}[1]
\REQUIRE $\data^s$ for test person $\hat{\person}$ obtained via micro-interactions, 
\REQUIRE $\theta$; Meta-model
\STATE Initialise $\hat{\theta} = \theta$
  \FOR{$i=0$ \TO $meta\_gs$}
    \STATE Evaluate $\nabla_{\hat{\theta}} \lossi(\learner_{\hat{\theta}})$ w.r.t. $\data^s$
    \STATE Compute adapted parameters with gradient descent: $\hat{\theta}'=\hat{\theta}-\alpha \nabla_{\hat{\theta}}  \lossi(  \learner_{\hat{\theta}} )$
  \ENDFOR
  \FORALL{$\data^q_i$}
    \STATE predict $y^q_i = \learner_{\hat{\theta}'} (\data^q_i)$
  \ENDFOR
\end{algorithmic}
\end{algorithm}
\end{minipage}

\subsection{Personalised RN}
\label{sec:prn}

Personalised Relation Networks~$RN^p$ learns a matching, generalisable to new persons encountered at test time. 
Design of person-tasks~($\person_i$) follow the methodology in Section~\ref{sec:pml} where the the support set, $\data^s$, and the query set $\data^q$, is selected from the same person.
Similar to $MAML^p$, we select $K^s$ amount of data instances per class to create the support set, $\data^s$ and we select all remaining data instances of the class to create the query set $\data^q$.
We create meta-training instance for $\person_i$ by combining each data instance $x^q_i$, in $\data^q$, with the support set, $\data^s$. 
Therefore, the matching is always performed against their own data in the support set. This method is similar to personalisation methods described for Personalised Matching Networks~\cite{wijekoon2020knowledge}.
At test time, a meta-test person $\hat{p}$ provide a support set, $\data^s$ with representatives for each class in $\classes$. This support set is thereafter combined with each test query instance $x^q_i$ for predicting class label $y^q_i$ using the $RN^p$ model as in Algorithm~\ref{algo:prn_test}.  
We present the training of Personalised RN in Algorithm~\ref{algo:prn_train}. 
A $RN^p$ has two parametric modules, one for feature representation learning, $\theta_f$ and one for similarity learning, $\theta_r$~(Figure~\ref{fig:rn}), and similar to $MAML^p$, feature representation learning module can be configured to suit heterogeneous sensor modalities or modality combinations.

\begin{minipage}{0.55\textwidth}
\begin{algorithm}[H]
\caption{Personalised RN Training}
\label{algo:prn_train}
\begin{algorithmic}[1]
\REQUIRE $p(\person)$: HAR dataset; distribution over persons
\REQUIRE $\alpha$: step size hyper-parameter
\STATE randomly initialise $\theta_f$ and $\theta_r$
\WHILE{not done}
    \STATE Sample a person $\person_i \sim p(\person)$
    \STATE $\data^s = \{(x,y) \in \person_i : |\data^s| = K^s \times |\classes| \}$
    \STATE $\data^q = \{(x,y) \in \person_i : |\data^q| = K^q \times |\classes|, \data^s \cap \data^q = \emptyset \}$
    \FORALL{$\data^q_i$}
        \STATE Create train data instance ($\data^q_i, \data^s$)
    \ENDFOR
    \STATE Evaluate $\nabla \lossi(\learnerfr)$ w.r.t. $K^q \times |\classes|$ number of train data instances
    \STATE Update $(\theta_f,\theta_r) \leftarrow (\theta_f,\theta_r) - \alpha \nabla_{\theta_f} \lossi(\learnerfr)$
\ENDWHILE
\end{algorithmic}
\end{algorithm}
\end{minipage}
\begin{minipage}{0.40\textwidth}
\begin{algorithm}[H]
\caption{Personalised RN Testing}
\label{algo:prn_test}
\begin{algorithmic}[1]
\REQUIRE Support set  $\data^s$ for test person $\hat{\person}$ obtained via micro-interactions, 
\REQUIRE $\theta_r, \theta_f$; Relation Network Model
  \FORALL{$\data^q_i$}
    \STATE predict $y^q_i = \learnerfr (\data^q_i,\data^s)$
  \ENDFOR
\end{algorithmic}
\end{algorithm}
\end{minipage}

\section{Comparative Study}
\label{sec:compare}
We compare the performance of Personalised Meta-learning algorithms, \textbf{Personalised MAML}~($MAML^p$) and \textbf{Personalised RN}~($RN^p$) against a number of baselines and the state-of-the-art algorithms as listed below;

\begin{description}
\item [DL:] Best performing DL algorithm from benchmark performances published in~\cite{wijekoon2019mex}

\item [MN:] Matching Networks from~\cite{vinyals2016matching}; Few-shot Learning classifier
\item [MAML:] Model-Agnostic Meta-Learner~\cite{finn2017model} (detailed in Section~\ref{sec:meta-learning}) The state-of-the-art for Few-shot Image classification
\item [RN:] Relation Networks~\cite{sung2018learning} (detailed in Section~\ref{sec:meta-learning}) State-of-the-art for Few-shot Image classification

\item [MN$^p$:] Personalised Matching Networks from~\cite{wijekoon2020knowledge}; Few-shot Learning classifier, state-of-the-art for personalised HAR

\item [MAML$^p$ (Ours):] Personalised MAML introduced in Section~\ref{sec:pmaml}
\item [RN$^p$ (Ours):] Personalised Relation Networks introduced in Section~\ref{sec:prn}
\end{description}

\subsection{Datasets and Pre-processing}
We use three data sources to create 9 datasets in single modality sensing.
Both $MAML^p$ and $RN^p$ are model agnostic, such that the feature representation learning models are interchangeable to suit any modality combination.

\mex~\footnote{https://archive.ics.uci.edu/ml/datasets/MEx} is a Physiotherapy Exercises dataset complied with 30 participants performing 7 exercises. A participant performs one exercise for only 60 seconds. A depth camera~(DC), a pressure mat~(PM) and two accelerometers on the wrist~(ACW) and the thigh~(ACT) provide four sensor data streams creating four datasets.
\pamap~\footnote{http://archive.ics.uci.edu/ml/datasets/pamap2+physical+activity+monitoring} dataset contains 8 Activities of Daily Living recorded with 8 participants. Three accelerometers on the hand~(H), the chest~(C) and the ankle~(A) provide three sensor data streams creating three datasets.
\sfb~\footnote{https://github.com/rgu-selfback/Datasets} is a HAR dataset with 6 ambulatory and 3 stationary activities. These activities are recorded with 33 participants using two accelerometers on the wrist~(W) and the thigh~(T), creating two datasets.

A sliding window method is applied on each sensor data stream to obtain data instances. Window size of 5 seconds is applied for all 9 datasets and an overlap of 3, 1 and 2.5 for data sources \mex, \pamap and \sfb, resulted in 30, 76 and 88 data instance per person-activity on average.
A few pre-processing steps are applied on data instances, specific to their sensor modalities. DC and PM modalities use a reduced frame rate from $15Hz$ to $1Hz$ and DC frame size is reduced from $240\times320$ to $12\times16$.
Accelerometer data apply DCT feature transformation on every 1 second data slice of each axis and select the 60 most prominent DCT coefficients.
Resulting input sizes for $\theta_f$ of RN and $\theta$ of MAML are $(5\times12\times16)$, $(5\times16\times16)$ and $(5\times3\times60)$ for DC, PM and AC modalities respectively.

\subsection{Implementation}

\begin{table}[ht]
\row{1.2}
\centering
\caption{Best performing DL Network Architectures of the 9 datasets, $td$:TimeDistributedLayer, $conv(k)n$:ConvolutionalLayer with $n$ kernels of kernel size $k$, $maxpool(k)$:MaxPoolingLayer with pool size $k$, $dense(k)$:DenseLayer with $k$ units, $bn$:BatchNormalisation}
\label{tbl:dls}
\begin{tabular} {p{3.5cm} p{9cm}}
\hline
Datasets&Architecture\\
\hline
\mexdc&$conv(3.3)32\rightarrow maxpool(2.2)\rightarrow bn \rightarrow conv(3.3)64\rightarrow maxpool(2.2)\rightarrow bn \rightarrow flatten \rightarrow dense(1200)\rightarrow bn \rightarrow dense(600)\rightarrow bn \rightarrow dense(100)\rightarrow bn $\\
\mexact, \mexacw, \mexpm, \pamapa, \pamapc, \pamaph, \sbt, \sbw&$td-conv(5)32\rightarrow maxpool(2)\rightarrow bn \rightarrow td-conv(5)64\rightarrow  maxpool(2)\rightarrow bn \rightarrow td-flatten \rightarrow lstm(1200)\rightarrow bn  \rightarrow dense(600)\rightarrow bn  \rightarrow dense(100)\rightarrow bn $\\
\hline
\end{tabular}
\end{table}

\begin{figure}[ht]
\centering
\begin{subfigure}{.25\textwidth}
\centering
\includegraphics[width=0.92\linewidth]{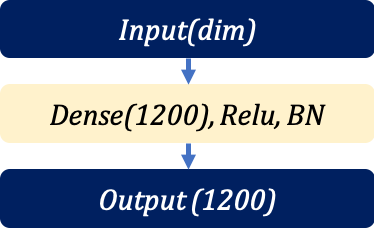}
\caption{$\theta$}
\label{fig:maml_theta}
\end{subfigure}%
\begin{subfigure}{.35\textwidth}
\centering
\includegraphics[width=0.92\linewidth]{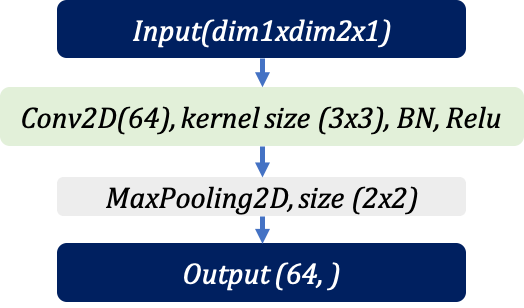}
\caption{$\theta_f$}
\label{fig:rn_f}
\end{subfigure}%
\begin{subfigure}{.35\textwidth}
\centering
\includegraphics[width=0.92\linewidth]{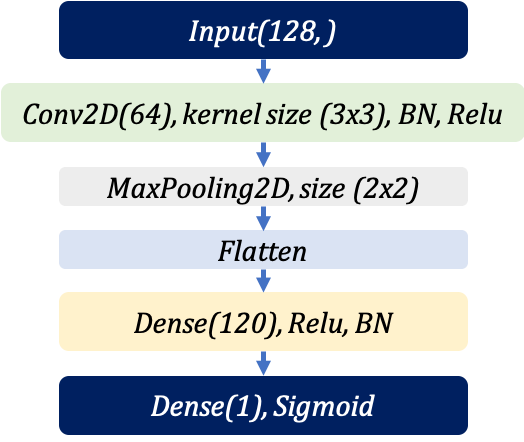}
\caption{$\theta_r$}
\label{fig:rn_r}
\end{subfigure}%
\caption{Network Architectures}
\label{fig:theta}
\end{figure}

DL benchmark results with the \mex datasets are published in~\cite{wijekoon2019mex} and for comparability we implement the same network architectures and evaluate with the best performing baselines for \pamap and \sfb datasets.
We detail the selected network architectures in Table~\ref{tbl:dls}.

We use the $\theta$ network in Figure~\ref{fig:maml_theta} as the feature representation learning model in $MN$, $MN^p$, $MAML$ and $MAML^p$ algorithms, where the input sizes for modalities Accelerometer, Depth camera and Pressure mat are reshaped to 900, 960 and 1280 respectively.
$RN$ and $RN^p$ use the network in Figure~\ref{fig:rn_f} to learn the feature representations, where the input sizes for modalities Accelerometer, Depth camera and Pressure mat are ($5 \times 180$), ($80 \times 12$) and ($80 \times 16$) respectively with 1 channel. Relation module $\theta_f$ uses the network in Figure~\ref{fig:rn_r} where the input is twice the size of the output of $\theta_f$ (See Figure~\ref{fig:rn} for complete network architecture).

We use the best performing hyper-parameter settings for $MN$ and $MN^p$ from~\cite{wijekoon2020knowledge} where the networks are 5-shot classifiers, trained for 20 epochs with early stopping using Categorical Cross-entropy as the objective.
We train $RN$ and $RN^p$ similar to $MN$, but using Mean Squared Error as the objective function~\cite{sung2018learning} where $\alpha = 0.001$, $K^s = 5$, trained for 300 epochs and apply early stopping.
Our initial empirical evaluations showed that using $RN$ and $RN^p$ trained using Categorical Cross-entropy yields comparable results and achieves model convergence faster, compared to using Mean Squared Error.
$MAML$ and $MAML^p$ are using Categorical Cross-entropy as the objective function and use $K^s = 5$, $gs = 5$ and $meta\_gs = 10$ for training and testing.
We use $\alpha = 0.4$ and $\beta = 0.001$, $n=32$ and is trained for 100 epochs.
All models are trained with the Adam optimiser and the Meta-Learning models do not use mini-batching.

\subsection{Evaluation Methodology}
We follow the person-aware evaluation methodology, Leave-One-Person-Out~(LOPO) in our experiments. We leave data from one person to create meta-test tasks and use the rest to create meta-train tasks.
We note that during testing, even $MN$, $MAML$ and $RN$ preserve the personalisation aspect because of the LOPO evaluation strategy where only one user is present in the meta-test tasks.
The meta-train and meta-test tasks are created while maintaining class balance; accordingly we report the accuracy of each experiment averaged over the number of person folds. LOPO evaluation methodology require a non-parametric statistical significance test as they produce results that are not normally distributed. We use the Wilcoxon signed-rank test for paired samples to evaluate the statistical significance at 95\% confidence and highlight the significantly improved performances in bold text.

\subsection{Results}
\label{sec:results}

\begin{table}[ht]
\caption{Personalised Meta-Learner performance comparison for Exercise Recognition}
\label{tbl:mex}
\row{1.2}
\centering
\begin{tabular} {l r r r r}
\hline
Algorithm&\mexact&\mexacw&\mexdc&\mexpm\\
\hline
DL&0.9015&0.6335&0.8720&0.7408\\
\hline
$MN$&0.9073&0.4620&0.5065&0.6187\\
$MN^p$&0.9155&0.6663&0.9342&0.8205\\
\hline
$MAML$&0.8673&0.6525&0.9629&0.9283\\
$RN$&0.9327&0.7279&0.8189&0.8145\\
\hline
$MAML^p$(Ours)&0.9106&0.6834&\textbf{0.9795}&\textbf{0.9408}\\
$RN^p$(Ours)&\textbf{0.9436}&\textbf{0.7719}&0.9205&0.8520\\
\hline
\end{tabular}
\end{table}

Table~\ref{tbl:mex} presents the comparison of performances obtained by the algorithms in Section~\ref{sec:compare} for the Exercise Recognition~(\exrec) task using the four datasets derived from \mex.
As expected personalised Meta-Learning models significantly outperformed conventional DL and Meta-Learning models in all four experiments.
Notably the two datasets with accelerometer data recorded best performance with $RN^p$ while datasets with visual data; \mexdc and \mexpm, recorded best performance with $MAML^p$. It is noteworthy that the Personalised Few-shot Learning algorithm $MN^p$, achieves comparable performance against $MAML^p$ model of the \mexact dataset and outperform $RN^p$ model of the \mexdc dataset. When comparing conventional Meta-Learners~(i.e. $RN$, $MAML$) and Personalised Few-Shot Learner $MN^p$, we highlight that, $MN^p$ models achieve comparable performances or significantly outperform at least one conventional Meta-Learner with all four experiments. These results further confirm the importance of personalisation for \exrec.

Table~\ref{tbl:pmpsb} presents results for Ambulatory, Stationary and ADL activities using 5 datasets for \pamap and \sfb.
Similar to \exrec, Personalised Meta-Learning models have significantly outperformed conventional DL models, with both ambulatory and stationary activity data.
Notably, two experiments with ADL data, have significantly outperformed DL models with at least one of the personalised Meta-Learner models~(\pamaph: $MAML^p$, and $RN^p$, \pamapc: $RN^p$).
However Personalised Meta-Learner models fail to outperform DL models using the \pamapa dataset.
All $MAML^p$ models significantly outperform its original counterpart $MAML$, and $RN^p$ significantly outperform $RN$ with four experiments with the exception of \pamaph where $RN^p$ performance is comparable with $RN$.
While two of the five experiments significantly outperform Personalised $MN^p$, three experiments fail to outperform $MN^p$. But all experiments achieve their best performance with a personalised algorithms further confirming the significance of Personalisation in different domains of HAR.
$MN^p$'s use of the simpler similarity metric (such as cosine) has proven to be sufficient for \pamap in particular, compared to the sophisticated similarity model learnt with $RN$.

\begin{table}[ht]
\caption{Personalised meta-Learner performance comparison for Ambulatory, Stationary and ADL Activity Recognition}
\label{tbl:pmpsb}
\row{1.2}
\centering
\begin{tabular} {l r r r r r}
\hline
Algorithm&\sbt&\sbw&\pamaph&\pamapc&\pamapa\\
\hline
Best&0.7880&0.6997&0.7505&0.7878&0.8075\\
\hline
MN&0.8392&0.7669&0.6625&0.7536&0.7361\\
$MN^p$&0.9124&\textbf{0.8653}&0.7484&\textbf{0.8548}&\textbf{0.8330}\\
\hline
MAML&0.8398&0.7532&0.7593&0.7626&0.6830\\
RN&0.9334&0.8276&0.7818&0.8170&0.7527\\
\hline
$MAML^p$(Ours)&0.8625&0.8075&\textbf{0.8037}&0.7822&0.7256\\
$RN^p$(Ours)&\textbf{0.9487}&0.8528&0.7868&0.8294&0.7761\\
\hline
\end{tabular}
\end{table}

Overall, considering all personalisation algorithms, we find that experiments with visual data prefer the optimisation based meta-learning algorithm~(i.e. $MAML^p$) and experiments with time-series data prefer learning to compare methods~(i.e. $MN^p$ and $RN^p$). It is noteworthy that $MAML^p$ and $MN^p$ use a 1-dense layer network~(Figure~\ref{fig:theta}) and $RN^p$ uses a 1-convolution layer network for feature representation learning while achieving significant performance improvements.
While meta-learning attributes to this improvement, it is further enhanced by the personalised learning strategies.
These results highlight that Meta-Learners and Personalisation are positively contributing towards eliminating the need for parametric models with many deep layers that require a large labelled data collection for training.
This is highly significant in the domain of HAR, where even a comprehensive data collection fails to cover all possible personal nuances and traits that a reasoning model may encounter in the future.

Evidently, model adaptation by re-training using a few data instances at test time, significantly improved meta-model performance for $MAML^p$, suggesting that learnt meta-model was transferable and obtaining an activity label for a given test query, is a simple inference task using the adapted model.
While $RN^P$ does not require model-retraining, obtaining the activity class label for a given query involves a more complex inference process; each data instance in the user provided support set and the query instance is converted to feature vectors and later concatenated~(as described in Section~\ref{sec:prn}) to obtain the relation scores and derive the activity class label.
We calculate the average time elapsed for obtaining an activity label on the \mexact query data instance, using both algorithms in a computer with 8GB RAM and 3.1 GHz Dual-Core processor.
While $MAML^p$ takes 0.0156 milliseconds for a single classification task, $RN^P$ takes 2.4982 milliseconds in a 1-shot setting and 3.7218 milliseconds in a 5-shot setting.
This is an important difference, when selecting an algorithm to operate in an edge device with limited resources, where a high response rate is necessary while maintaining performance.

\section{Conventional vs. Personalised Meta-Learners}
\label{sec:pvsnp}

Here we look closer at training aspects to understand how personalisation improves performance of Meta-Learners using \mexpm dataset.
We select \mexpm because it is representative of a few-shot learning HAR dataset with only 30 data instances for each ``person-activity'' class. 

\subsection{MAML vs. MAML$^p$}
We first investigate the performance improvements achieved by $MAML^p$ after using the Personalised Meta-Learning methodology against $MAML$. 
Accordingly we compare three algorithms, $MAML$ where meta-train and test tasks are created disregarding any person identifiers; $MAML^p$, as described in Section~\ref{sec:pmaml}; and Person-aware $MAML$. 
Here Person-aware $MAML$ can be seen as a lazy personalisation of MAML where a meta-train task is comprised of data instances selected from a set of persons. 
The support set contains, $K^s$, data instances for each activity class, where data for any given activity must be obtained from a single person, but different activity classes may obtain data from different persons. 
The query set will have data from a single person who may not have been selected to form the support set. 
This method still preserve the concept of ``person-activity'' only at the class label level, but not over the entire support set level. 
We visualise the impact of model adaptation at test time
using these different algorithms in both, $K^s=1$, and $K^s=5$, settings on the \mexpm dataset.

\begin{figure}[ht]
\centering
\includegraphics[width=0.65\linewidth]{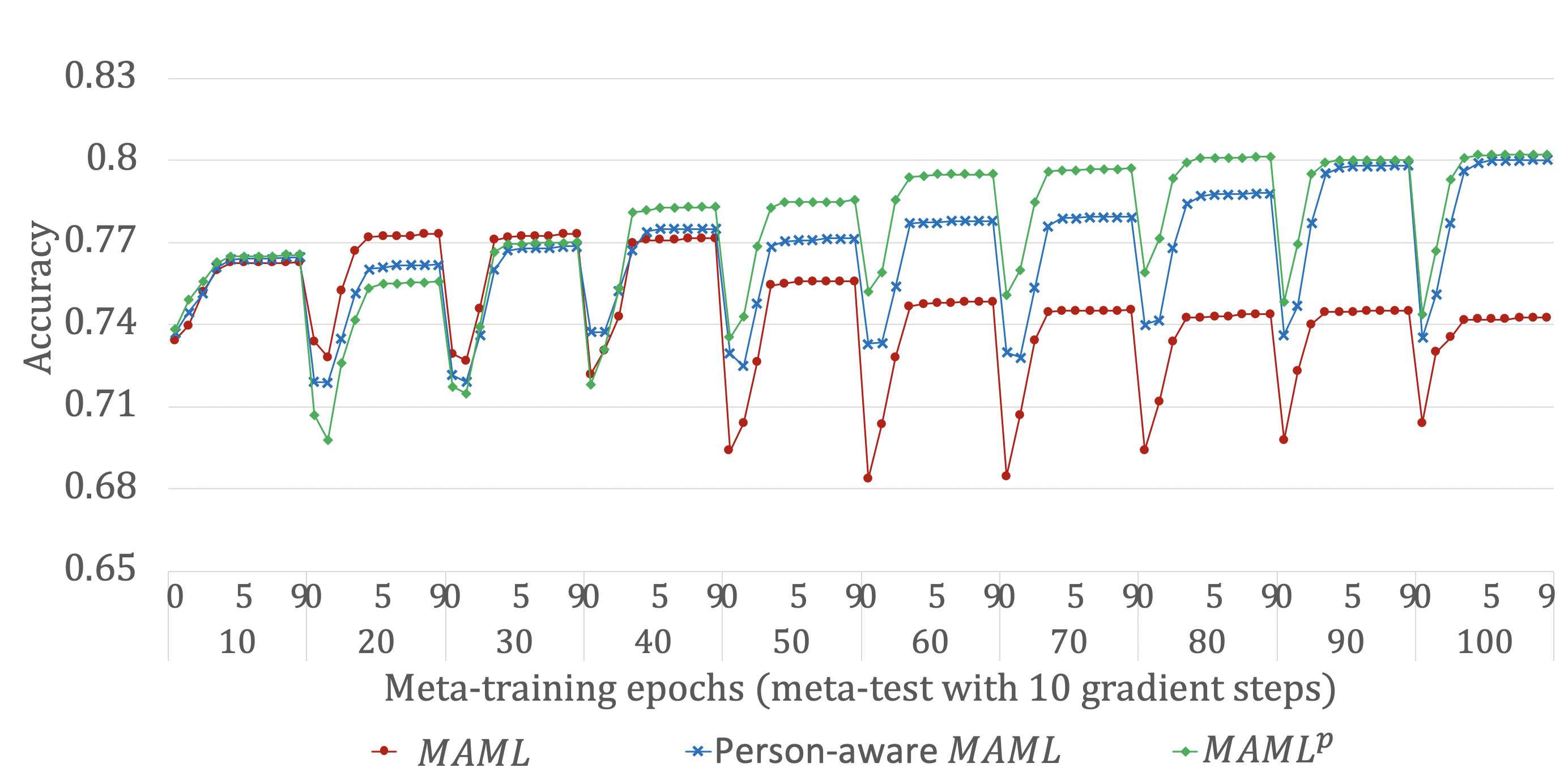}
\caption{MAML vs. Person-aware MAML vs. MAML$^p$ with \mexpm when $K^s = 1$}
\label{fig:maml_vs_mamlp_1shot}
\end{figure}
\begin{figure}[ht]
\centering
\includegraphics[width=.65\linewidth]{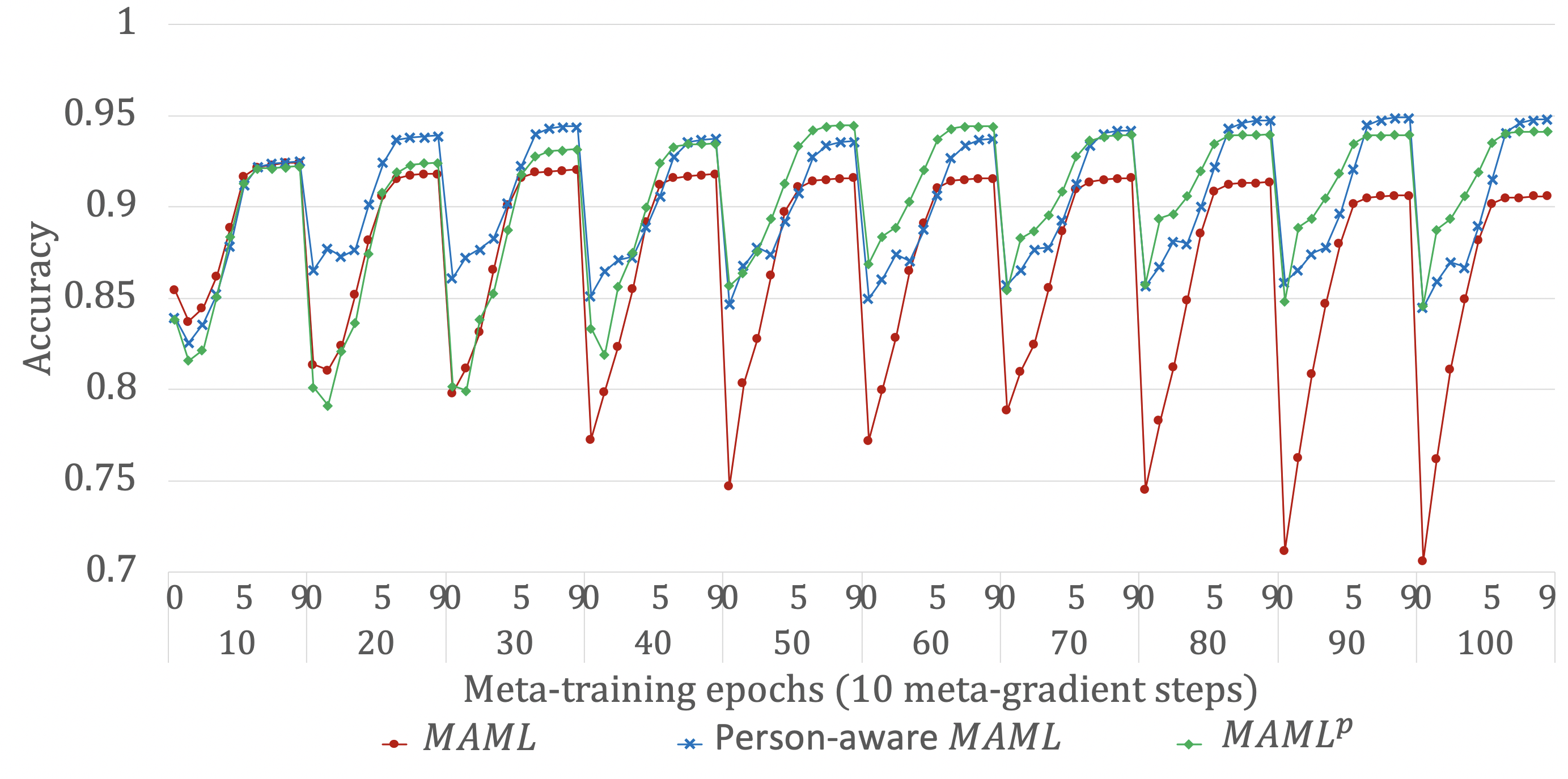}
\caption{MAML vs. Person-aware MAML vs. MAML$^p$ with \mexpm when $K^s = 5$}
\label{fig:maml_vs_mamlp_5shot}
\end{figure}
Figures~\ref{fig:maml_vs_mamlp_1shot} and~\ref{fig:maml_vs_mamlp_5shot} compare $MAML$, $MAML^p$ and Person-aware $MAML$ using \mexpm in $K^s=1$ and $K^s=5$ settings respectively.
Here we plot test-person accuracy (y-axis) evaluated at every 10 meta-train epochs~(2$^{nd}$ row of the x-axis); at each of these evaluation points, the meta-test support set is used to adapt the current meta-model for a further 10 meta-gradient steps~(1$^{st}$ row of the x-axis). 
During the adaptation steps we also record accuracy using the meta-test query.
Through this process we can observe the impact of the partially optimised general meta-model when being adapted for personalisation at test time (or at deployment) at different stages of optimisation.
$MAML^p$ and Person-aware $MAML$ significantly outperformed $MAML$ in both settings. When comparing $MAML^p$ and Person-aware $MAML$, $MAML^p$ algorithm achieves a more generalised meta-model even without performing meta-gradient steps for meta-model adaptation~(0 on $1^{st}$ row of the x-axis); this is most significant in the $K^s=1$ setting. 
These observations verify the advantage of creating personalised tasks; that even with the Person-aware $MAML$ algorithm, where each task contains data from multiple people, but each ``person-activity'' class only contains data from one person has clear benefits. 
Accordingly, personalised algorithms ensure that the task-models are trained for a set of ``person-activity'' classes instead of ``activity'' classes. 
$MAML^p$ where all ``person-activity'' data belongs to the same person, provides further generalisation with rapid adaptation.
Another indication of the significance of personalisation is found when investigating $MAML$ performance over the training epochs. While $MAML$ improves overall performance as the meta-model train, $MAML$ meta-test accuracy before adaptation~(every $0^{th}$ meta-gradient step), declines consistently. This is most significant when $K^s=5$, which indicates that the meta-model learned with $MAML$ is not generalised when an activity class in a meta-train task support set contains data from multiple people. In comparison, meta-model learned with $MAML^p$, performs well on meta-test tasks, even before adaptation.

\subsection{RN vs. RN$^p$}

Similarly we compare the performance between the two algorithms Relation Networks~(RN) and Personalised RN~($RN^p$) to understand the effect of personalisation on network training and testing. 
For this purpose we train the two algorithms, with the \mexpm dataset in two setting $K^s=1$ and $K^s=5$ for 300 epochs, and evaluate the model at every 10 epochs using meta-test tasks.

\begin{figure}[ht]
\centering
\begin{subfigure}{.5\textwidth}
\centering
\includegraphics[width=.9\linewidth]{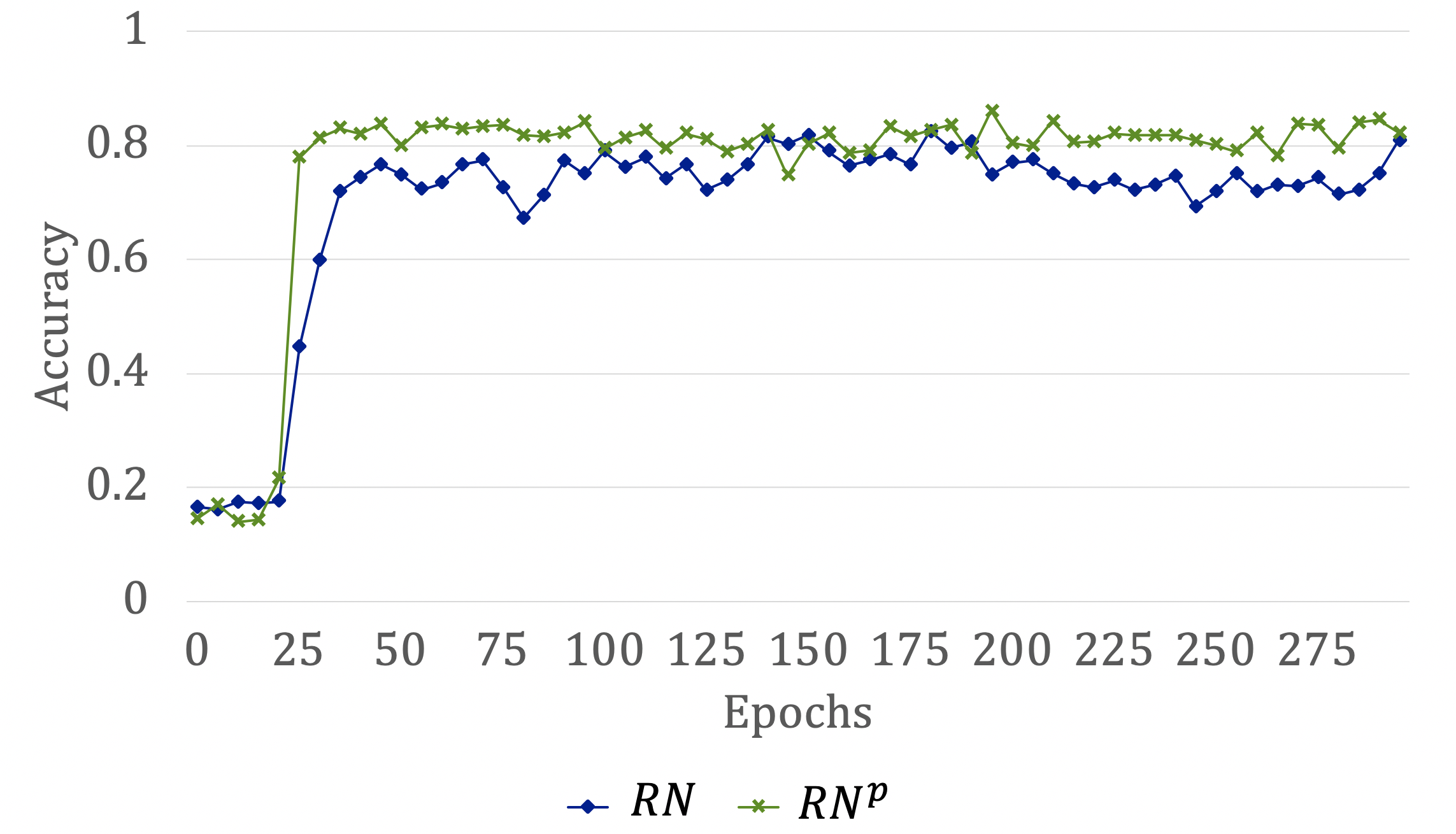}
\caption{$K^s=1$}
\label{fig:rn_pvsnp_1shot}
\end{subfigure}%
\begin{subfigure}{.5\textwidth}
\centering
\includegraphics[width=.9\linewidth]{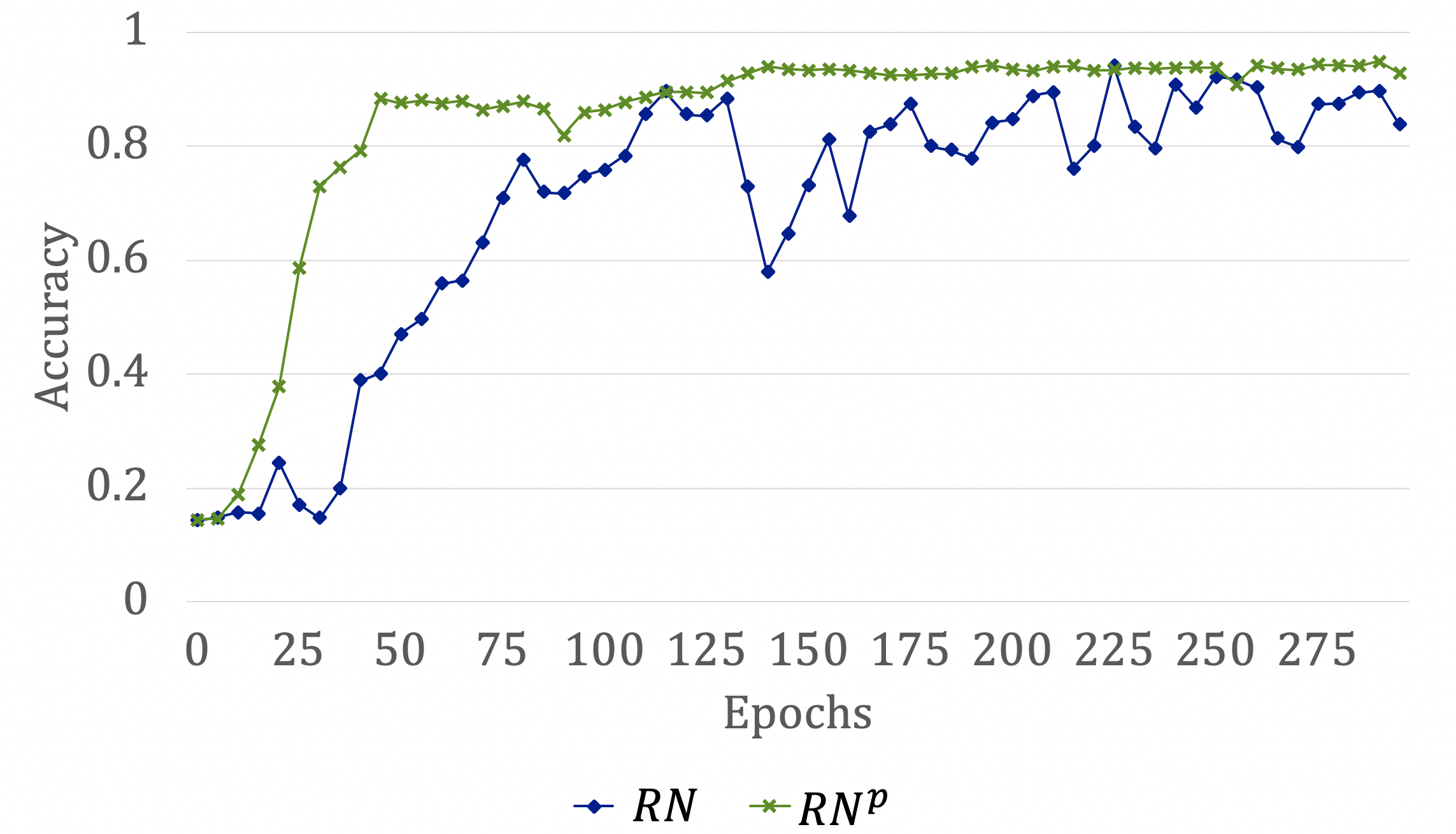}
\caption{$K^s=5$}
\label{fig:rn_pvsnp_5shot}
\end{subfigure}
\caption{RN vs. RN$^p$ with \mexpm meta-model tested at every 10 meta-train epochs}
\label{fig:rn_pvsnp}
\end{figure}

In Figures~\ref{fig:rn_pvsnp_1shot} and~\ref{fig:rn_pvsnp_5shot} plot the test accuracy on meta-test tasks obtained at every 10 meta-train epochs for the two algorithms RN and $RN^p$ in the two settings $K^s=1$ and $K^s=5$. 
It is evident that personalisation has stabilised the meta-training process, where meta-test model performs consistently better with $RN^p$ models.
In contrast meta-test evaluation on the $RN$ models is erratic, especially evident when $K^s=5$. When training $RN$ in the $K^s=5$ setting, a task is created by disregarding  the person parameter, as a result, an activity class contains data instances from more than one person and learning similarities to many people has adversely affected the learning of the $RN$ meta-model.
Similarly, in the $K^s=1$ setting, when a task contains only one data instance per class, learning from ones own data with $RN^p$ is advantages in comparison to $RN$ where the data instance for a class is from one person but not strictly similar to query person. 

\section{Personalised Meta-Learner Hyper-parameter selection}
\label{sec:hyperpara}
We explore three hyper-parameters of Personalised Meta-Learners using the 4 datasets from \mex for Exercise Recognition. The 4 \mex datasets give us the opportunity to compare how different modalities recorded for the same set of activities respond in different hyper-parameter settings when adapting to new unseen persons. 

\subsection{Meta-train query set size comparison for MAML$^p$}

First we explore the most effective $K^q$ value for training $MAML^p$. 
Originally, $MAML$ experiments used $K^q=K^s$ for image classification~\cite{finn2017model}.
As shown in Algorithm~\ref{algo:pmaml_train} line 8, $K^q$ determine how many data instances are considered in meta-update loss calculation, that later affect the meta-model learning. 
Given each meta-train task for $MAML^p$, now belong to a specific person, we expect to find the effect of using a fewer or larger number of data instance in meta-task evaluation. 
Accordingly we explore three $K^q$ values 5, 10 and $K-K^s$ where the $K$ is all available data instances for an ``person-activity'' class~(30 on average for all \mex datasets)

\begin{figure}[ht]
\centering
\begin{subfigure}{.5\textwidth}
\centering
\includegraphics[width=.7\linewidth]{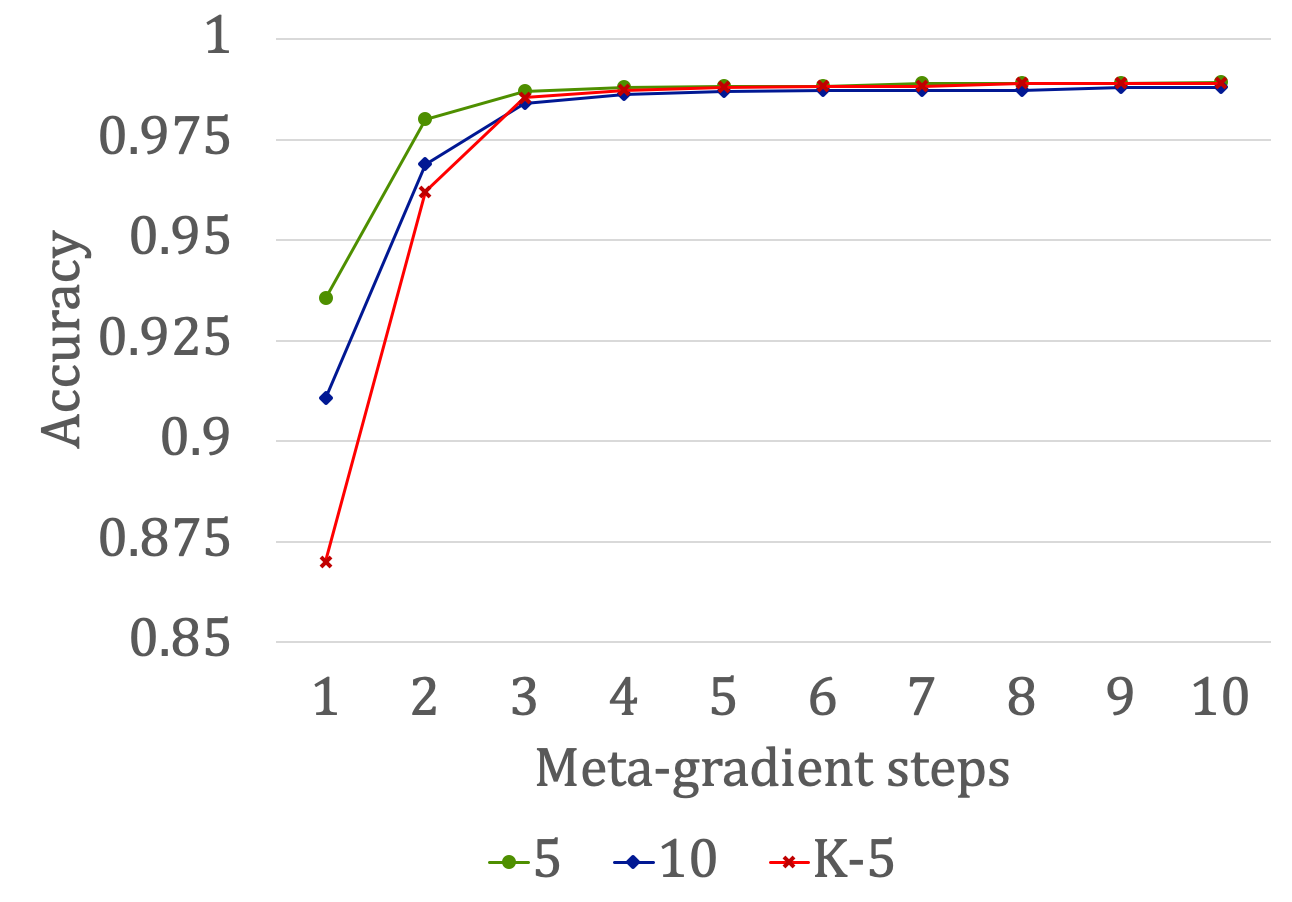}
\caption{PM}
\label{fig:maml_kq_pm}
\end{subfigure}%
\begin{subfigure}{.5\textwidth}
\centering
\includegraphics[width=.7\linewidth]{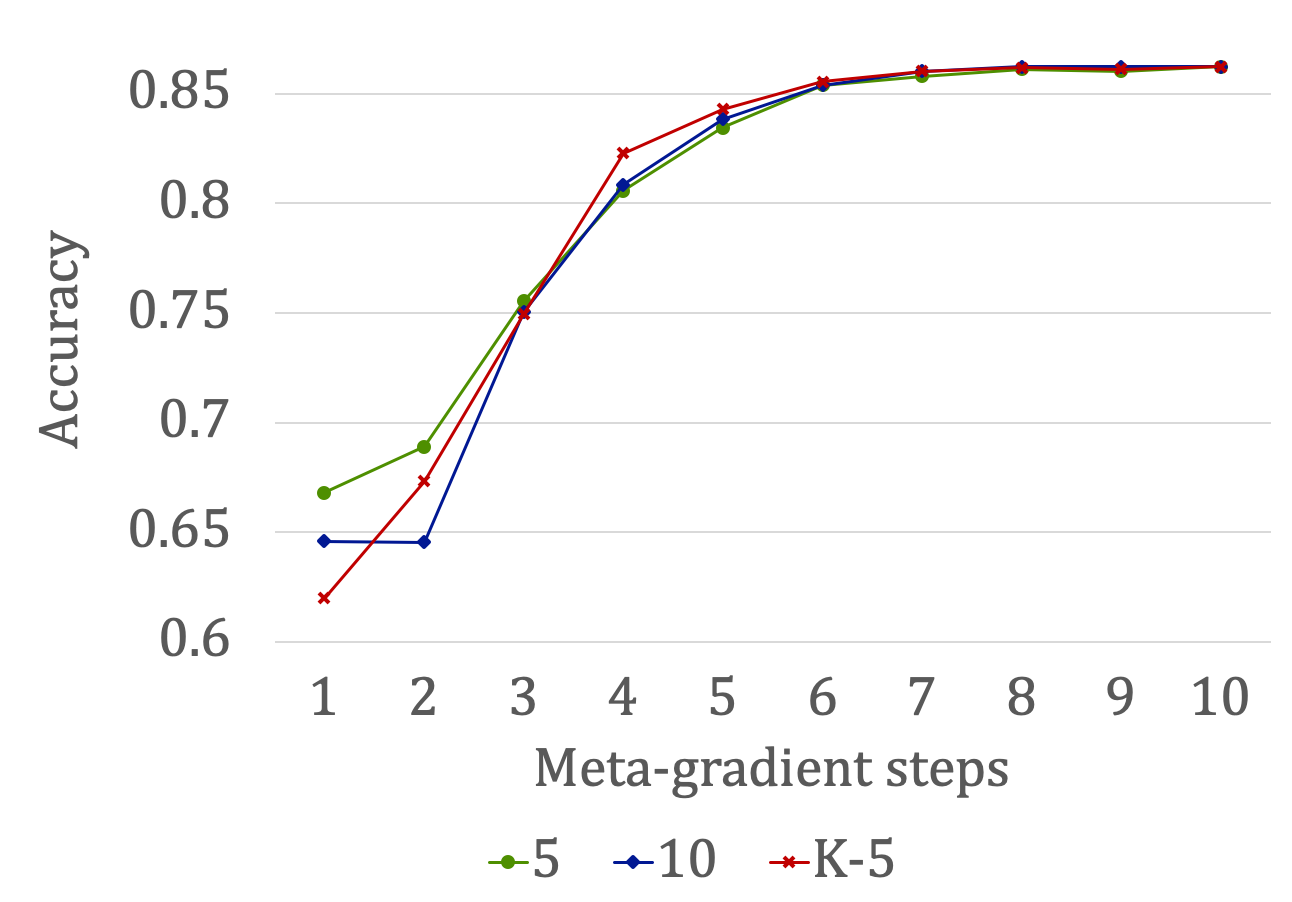}
\caption{ACT}
\label{fig:maml_kq_act}
\end{subfigure}
\caption{MAML$^p$: Meta-model adaptation with meta-models trained with different $K^q$ sizes}
\label{fig:maml_kq}
\end{figure}

Both settings, $K^s=1$ and $K^s=5$ achieved comparable performances with all three $K^q$ values. This suggests that meta-update is not affected adversely by creating personalised tasks, where the meta-update uses the loss computed from a batch of person-specific task models. 
We plot the meta-test performance during the 10 meta-gradient steps of meta-model adaptation to visualise the effect of learning a meta-model from different $K^q$ values in Figure~\ref{fig:maml_kq} (Here $K^s$ is set to 5). 
Both \mexact and \mexpm achieve similar performances with all three meta-models post-adaptation, but the most generalised meta-model is learned when using $K^q=5$. This is more significantly seen with the \mexpm experiment where $K^q = 5$ outperform other variants before and during model adaptation~(before: 6.55\% difference with \mexpm, 4.8\% difference with \mexact). 
These results suggest, limiting $K^q$ in each meta-train task improves generalisation of the meta-model for later adaptation with persons not seen during model training. 

\subsection{Support set size comparison for MAML$^p$}
Next we perform an exploratory study of $K^s$ using a range of $K^s$ values. Finding the the balance between $K^s$ and performance of $MAML^p$ is important because, at deployment, the test-person is expected to provide $K^s$ amount of data instances per activity class, therefore its is desirable to keep $K^s$ to a required minimum. We perform experiments with all 4 datasets from \mex for $K^s$ values 1, 3, 5, 7 and 10. 

\begin{figure}
\centering
\begin{subfigure}{.55\textwidth}
\centering
\includegraphics[width=0.8\linewidth]{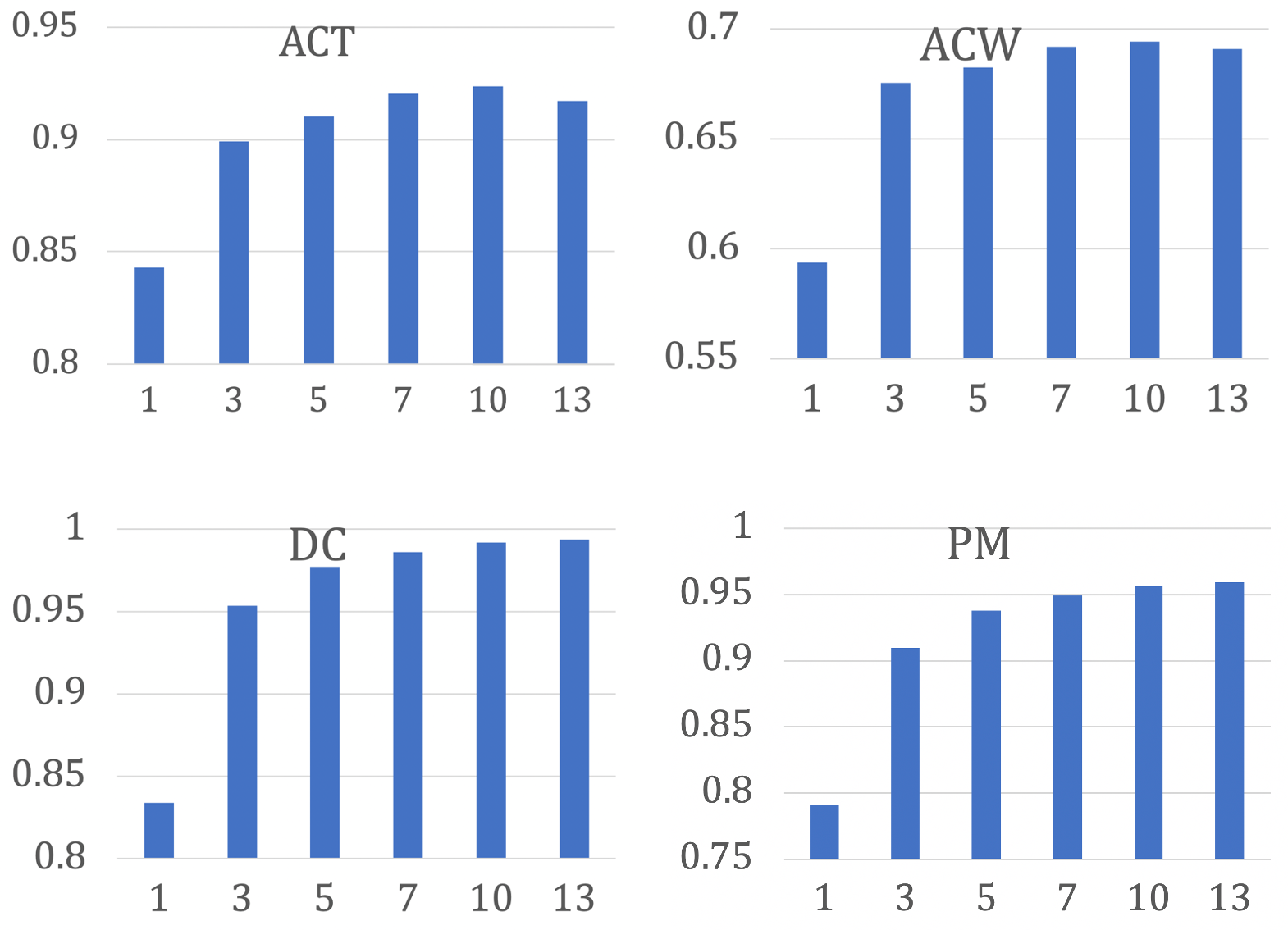}
\caption{Results of different $K^s$ with each \exrec dataset }
\label{fig:maml_kshot}
\end{subfigure}%
\begin{subfigure}{.38\textwidth}
\centering
\includegraphics[width=1\linewidth]{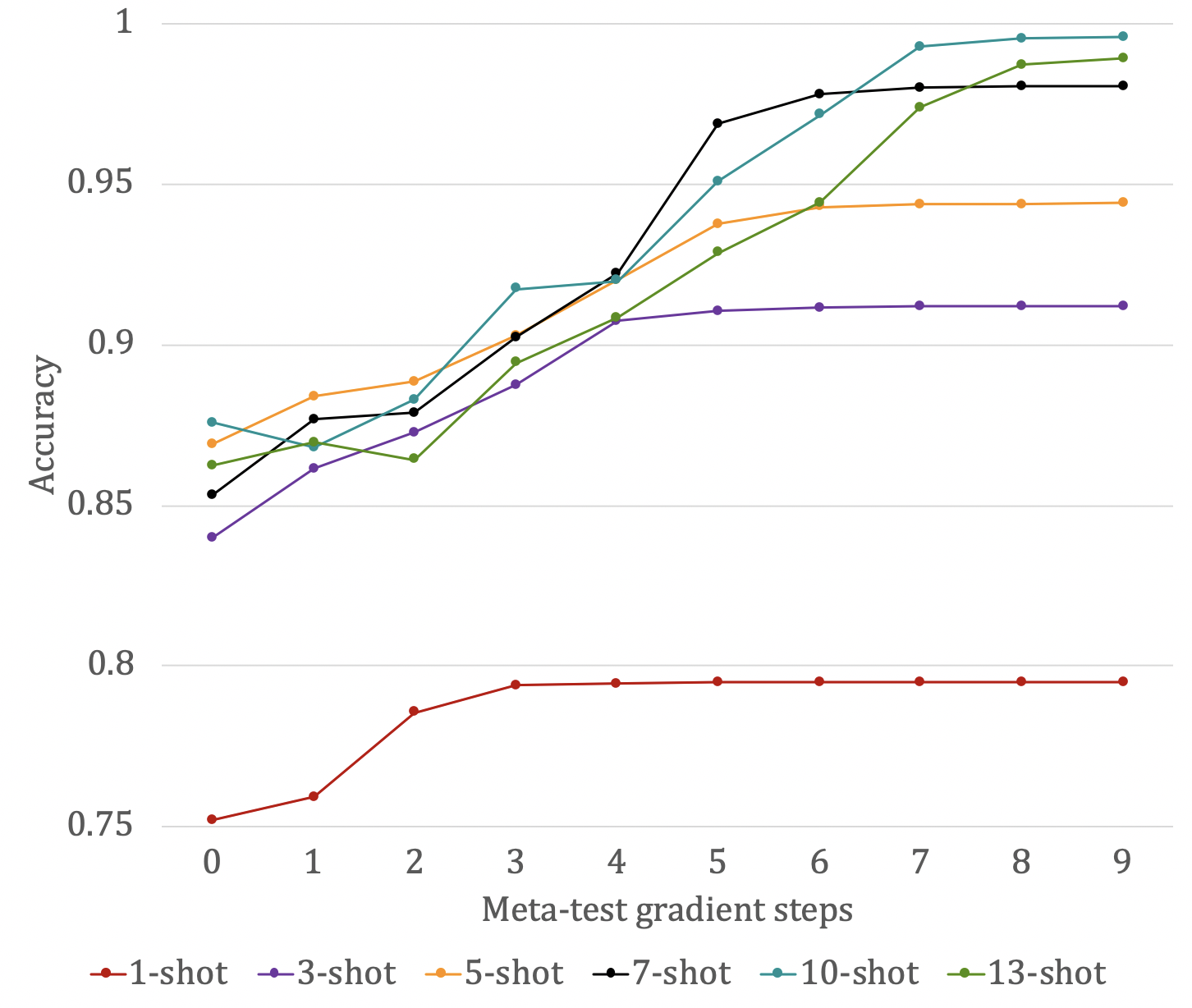}
\caption{Meta-model adaptation with meta-models trained with different $K^s$ values using \mexpm dataset}
\label{fig:maml_pm_kshot_test}
\end{subfigure}
\caption{MAML$^p$: Exploring meta-training with different $K^s$}
\label{fig:maml_kshot_train}
\end{figure}

Figure~\ref{fig:maml_kshot} plot meta-test performances obtained by datasets \mexact, \mexacw, \mexdc and \mexpm. Increasing $K^s$ consistently improve performance up to $K^s=10$ and report decreased or similar performance  when $K^s=13$ with all datasets.
A significant performance improvement is observed when increasing $K^s$ from 1 to 3. While modalities ACT and ACW achieve highest performance with $K^s=10$ modalities DC and PM gradually improve performance even at $K^s=13$. The meta-test accuracy at every meta-gradient update with increasing $K^s$ for \mexpm and \mexact for a randomly selected person is visualised in Figure~\ref{fig:maml_pm_kshot_test}. These figures indicate that meta-gradient adaptation converge faster when $K^s$ is smaller. Overall they validate that that increasing $K^s$ improve meta-test performance, before, during and after adaptation, most significantly seen when increasing $K^s$ from 1 to 3.

\subsection{Support set size comparison for RN$^p$}

Finally we explore different $K^s$ values to find the most optimal value for $RN^p$. Similar to $MAML^p$, $K^s$ determine how many data instances are required from a new person for the optimal personalisation of the model; therefore, we try to find the balance between performance improvement and $K^s$. Accordingly we create 10 experiments with each \mex dataset where $K^s$ range from 1 to 10. 

\begin{figure}[ht]
\centering
\includegraphics[width=0.6\textwidth]{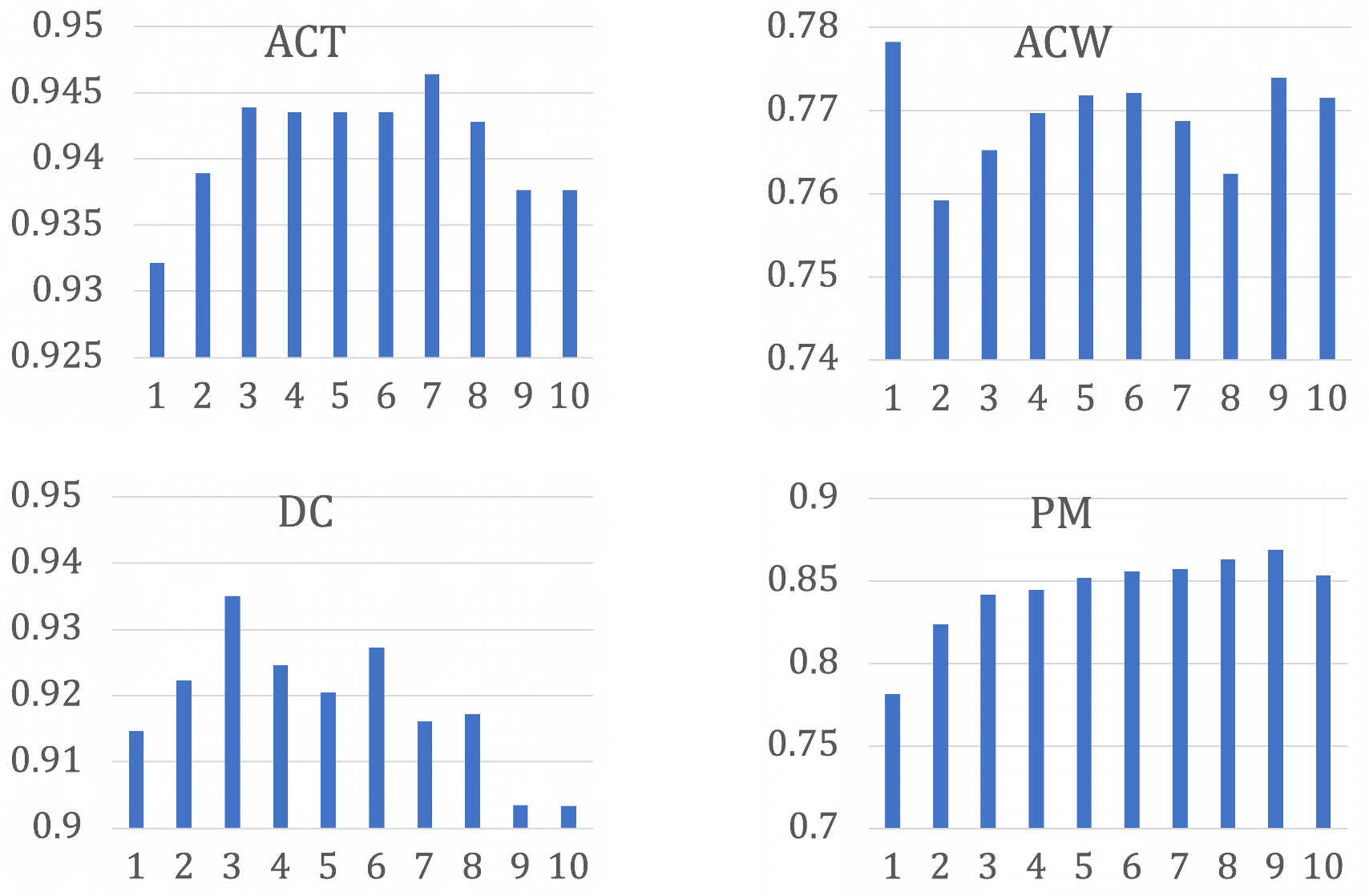}
\caption{RN$^p$: Exploring meta-training with different $K^s$ for each \exrec dataset}
\label{fig:rn_ks}
\end{figure}

We plot accuracy against different $K^s$ settings for each dataset in Figure~\ref{fig:rn_ks}. The figure clearly indicate that different sensor modalities prefer different $K^s$ values.
Both \mexact and \mexdc exhibit dome shaped behaviour in increasing $K^s$ settings where after $K^s=7$ it is detrimental to the network to have many examples of the same activity class for comparison. 
In contrast, \mexpm dataset show continuous performance improvement with increasing $K^s$ and at $K^s=10$ it starts to dome like \mexact and \mexdc modalities. 
We note that \mexacw behave differently to others when increasing $K^s$, while there is a domed behaviour for $K^s$ from 2 to 8, performances when $K^s = 1,9$ and $10$ are outliers that we will further investigate in future. 

As mentioned in Section~\ref{sec:results}, with the $RN^p$ algorithm, a larger $K^s$ value not only increase the amount of data instances requested from a test person, but also increases the memory and computational requirements.  Larger $K^s$ settings increases the number of comparisons and takes longer to perform a single classification task. Overall, we find $K^s = 5$ exhibit a proper balance between performance vs. memory requirements.

\section{Discussion}
\label{sec:discuss}

The comparative results from Section~\ref{sec:results} show that while $RN^p$ achieve best performance with many HAR datasets, the response rate is 248 times slower compared to $MAML^p$. An HAR algorithm should be able to recognise activities as they are performed in real-time for the best user-experience, and the processor and memory requirements along with the response time is crucial considerations for edge device deployment. In comparison, $MAML^p$ require post-deployment model re-training, which require the algorithm to perform in a development friendly environment using libraries like TensorFlow Light or PyTorch Mobile. 

A limitation of Personalised MAML and MAML in general is the inability to perform open-ended HAR. Both $MAML$ and $RN$ perform Zero-shot Learning for image classification~\cite{finn2017model,sung2018learning} for a fixed class length. Specifically, $MAML$ is restricted to performing multi-class classification with a conventional soft-max layer; for instance 5 outputs for a 5-class~(5-way) classification task. Open-ended HAR require dynamic expansion of the decision layer as the person add new activities in addition to the activities that are already included. Few-shot classifiers such as Matching Networks (MN)~\cite{vinyals2016matching} does not have a strict decision layer which inspired Open-ended MN~\cite{wijekoon2020knowledge} for Open-ended HAR. Similarities of Relation Networks~(RN) to MN presents the opportunity to improve Open-ended HAR using RN, which we will explore in future.  

When a Personalised Meta-Learning model is trained and embedded in the fitness application, there is a initial configuration step that is required for collecting the calibration data(i.e. support set) of the end-user.
The end-user will be instructed to record a few seconds of data for each activity using the sensor modalities synchronised with the fitness application. 
This is similar to demographic configurations users perform when installing new fitness applications~(on-boarding). 
Thereafter this support set will be used by the algorithm either to re-train the model~($MAML^p$) or for comparison~($RN^p$). 
Both $MAML^P$ and $RN^p$ provides the opportunity to provide new calibration data if the physiology of the user change to improve performance. Such changes include a gait change, a disability or a dramatic weight change that affect their personal activity rhythms. 
\section{Conclusion}
\label{sec:conc}
In this paper, we presented Personalised Meta-Learning, a methodology for learning to optimise for personalisation of Human Activity Recognition~(HAR) using only q few labelled data.
This is achieved by treating the ''person-activity'' pair in a HAR dataset as a class label, where each class now only has few instances of data for training.
Accordingly, we implement Personalised Meta-Learning with two Meta-Learning algorithms for few-shot classification Personalised MAML~( $MAML^p$) and Personalised Relation Networks~($RN^p$) where a meta-model is learned, such that it can be rapidly adapted to any person not seen during training.
Both algorithms require only a few instances of calibration data from the end-user to personalised the meta-model, where at deployment, $MAML^p$ uses calibration data for adaptation with model re-training and $RN^p$ uses calibration data directly for matching (without re-training).
Our evaluation with 9 HAR datasets shows that both algorithms achieve significant performance improvements in a range of HAR domains while outperforming conventional Deep Learning, Few-shot Learning and Meta-Learning algorithms. We highlight that  personalisation achieves higher model generalisation, compared to non-personalised Meta-Learners, which results is faster model adaptation.
Importantly we find, while $RN^p$ outperform $MAML^p$ with a majority of HAR datasets, $MAML^p$ performs is significantly faster than $RN^p$ due to the gains over paired matching.

\bibliography{ref}

\begin{thebibliography}{18}
\providecommand{\natexlab}[1]{#1}
\providecommand{\url}[1]{\texttt{#1}}
\expandafter\ifx\csname urlstyle\endcsname\relax
  \providecommand{\doi}[1]{doi: #1}\else
  \providecommand{\doi}{doi: \begingroup \urlstyle{rm}\Url}\fi

\bibitem[Berchtold et~al.(2010)Berchtold, Budde, Gordon, Schmidtke, and
  Beigl]{berchtold2010actiserv}
Martin Berchtold, Matthias Budde, Dawud Gordon, Hedda~R Schmidtke, and Michael
  Beigl.
\newblock Actiserv: Activity recognition service for mobile phones.
\newblock In \emph{International Symposium on Wearable Computers (ISWC) 2010},
  pp.\  1--8. IEEE, 2010.

\bibitem[Finn et~al.(2017)Finn, Abbeel, and Levine]{finn2017model}
Chelsea Finn, Pieter Abbeel, and Sergey Levine.
\newblock Model-agnostic meta-learning for fast adaptation of deep networks.
\newblock In \emph{Proceedings of the 34th International Conference on Machine
  Learning-Volume 70}, pp.\  1126--1135. JMLR. org, 2017.

\bibitem[Graves et~al.(2014)Graves, Wayne, and Danihelka]{graves2014neural}
Alex Graves, Greg Wayne, and Ivo Danihelka.
\newblock Neural turing machines.
\newblock \emph{arXiv preprint arXiv:1410.5401}, 2014.

\bibitem[Longstaff et~al.(2010)Longstaff, Reddy, and
  Estrin]{longstaff2010improving}
Brent Longstaff, Sasank Reddy, and Deborah Estrin.
\newblock Improving activity classification for health applications on mobile
  devices using active and semi-supervised learning.
\newblock In \emph{2010 4th International Conference on Pervasive Computing
  Technologies for Healthcare}, pp.\  1--7. IEEE, 2010.

\bibitem[Mishra et~al.(2017)Mishra, Rohaninejad, Chen, and
  Abbeel]{mishra2017simple}
Nikhil Mishra, Mostafa Rohaninejad, Xi~Chen, and Pieter Abbeel.
\newblock A simple neural attentive meta-learner.
\newblock \emph{arXiv preprint arXiv:1707.03141}, 2017.

\bibitem[Miu et~al.(2015)Miu, Missier, and Pl{\"o}tz]{miu2015bootstrapping}
Tudor Miu, Paolo Missier, and Thomas Pl{\"o}tz.
\newblock Bootstrapping personalised human activity recognition models using
  online active learning.
\newblock In \emph{2015 IEEE International Conference on Computer and
  Information Technology; Ubiquitous Computing and Communications; Dependable,
  Autonomic and Secure Computing; Pervasive Intelligence and Computing}, pp.\
  1138--1147. IEEE, 2015.

\bibitem[Nichol et~al.(2018)Nichol, Achiam, and Schulman]{nichol2018first}
Alex Nichol, Joshua Achiam, and John Schulman.
\newblock On first-order meta-learning algorithms.
\newblock \emph{arXiv preprint arXiv:1803.02999}, 2018.

\bibitem[Ord{\'o}{\~n}ez \& Roggen(2016)Ord{\'o}{\~n}ez and
  Roggen]{ordonez2016deep}
Francisco~Javier Ord{\'o}{\~n}ez and Daniel Roggen.
\newblock Deep convolutional and lstm recurrent neural networks for multimodal
  wearable activity recognition.
\newblock \emph{Sensors}, 16\penalty0 (1):\penalty0 115, 2016.

\bibitem[Santoro et~al.(2016)Santoro, Bartunov, Botvinick, Wierstra, and
  Lillicrap]{santoro2016meta}
Adam Santoro, Sergey Bartunov, Matthew Botvinick, Daan Wierstra, and Timothy
  Lillicrap.
\newblock Meta-learning with memory-augmented neural networks.
\newblock In \emph{International conference on machine learning}, pp.\
  1842--1850, 2016.

\bibitem[Snell et~al.(2017)Snell, Swersky, and Zemel]{snell2017prototypical}
Jake Snell, Kevin Swersky, and Richard Zemel.
\newblock Prototypical networks for few-shot learning.
\newblock In \emph{Advances in neural information processing systems}, pp.\
  4077--4087, 2017.

\bibitem[Sun et~al.(2012)Sun, Kashima, and Ueda]{sun2012large}
Xu~Sun, Hisashi Kashima, and Naonori Ueda.
\newblock Large-scale personalized human activity recognition using online
  multitask learning.
\newblock \emph{IEEE Transactions on Knowledge and Data Engineering},
  25\penalty0 (11):\penalty0 2551--2563, 2012.

\bibitem[Sung et~al.(2018)Sung, Yang, Zhang, Xiang, Torr, and
  Hospedales]{sung2018learning}
Flood Sung, Yongxin Yang, Li~Zhang, Tao Xiang, Philip~HS Torr, and Timothy~M
  Hospedales.
\newblock Learning to compare: Relation network for few-shot learning.
\newblock In \emph{Proceedings of the IEEE Conference on Computer Vision and
  Pattern Recognition}, pp.\  1199--1208, 2018.

\bibitem[Tapia et~al.(2007)Tapia, Intille, Haskell, Larson, Wright, King, and
  Friedman]{tapia2007real}
Emmanuel~Munguia Tapia, Stephen~S Intille, William Haskell, Kent Larson, Julie
  Wright, Abby King, and Robert Friedman.
\newblock Real-time recognition of physical activities and their intensities
  using wireless accelerometers and a heart rate monitor.
\newblock In \emph{2007 11th IEEE international symposium on wearable
  computers}, pp.\  37--40. IEEE, 2007.

\bibitem[Vinyals et~al.(2016)Vinyals, Blundell, Lillicrap, Wierstra,
  et~al.]{vinyals2016matching}
Oriol Vinyals, Charles Blundell, Timothy Lillicrap, Daan Wierstra, et~al.
\newblock Matching networks for one shot learning.
\newblock In \emph{Advances in neural information processing systems}, pp.\
  3630--3638, 2016.

\bibitem[Wang et~al.(2019)Wang, Chen, Hao, Peng, and Hu]{wang2019deep}
Jindong Wang, Yiqiang Chen, Shuji Hao, Xiaohui Peng, and Lisha Hu.
\newblock Deep learning for sensor-based activity recognition: A survey.
\newblock \emph{Pattern Recognition Letters}, 119:\penalty0 3--11, 2019.

\bibitem[Wijekoon et~al.(2019)Wijekoon, Wiratunga, and Cooper]{wijekoon2019mex}
Anjana Wijekoon, Nirmalie Wiratunga, and Kay Cooper.
\newblock Mex: Multi-modal exercises dataset for human activity recognition.
\newblock \emph{arXiv preprint arXiv:1908.08992}, 2019.

\bibitem[Wijekoon et~al.(2020)Wijekoon, Wiratunga, Sani, and
  Cooper]{wijekoon2020knowledge}
Anjana Wijekoon, Nirmalie Wiratunga, Sadiq Sani, and Kay Cooper.
\newblock A knowledge-light approach to personalised and open-ended human
  activity recognition.
\newblock \emph{Knowledge-based systems}, 192:\penalty0 105651, 2020.

\bibitem[Yao et~al.(2017)Yao, Hu, Zhao, Zhang, and
  Abdelzaher]{yao2017deepsense}
Shuochao Yao, Shaohan Hu, Yiran Zhao, Aston Zhang, and Tarek Abdelzaher.
\newblock Deepsense: A unified deep learning framework for time-series mobile
  sensing data processing.
\newblock In \emph{Proceedings of the 26th International Conference on World
  Wide Web}, pp.\  351--360, 2017.

\end{thebibliography}
\bibliographystyle{iclr2020_conference}

\end{document}